\pdfoutput=1

\documentclass[10pt, a4paper, twocolumn, teaser, showabstract]{naverlabseurope}

\usepackage{times}
\usepackage{latexsym}
\usepackage{makecell}
\usepackage[utf8]{inputenc}
\usepackage[most]{tcolorbox}
\usepackage{xcolor}
\usepackage{caption}
\usepackage{subcaption}
\usepackage{mwe}
\usepackage{amsmath}
\usepackage{amssymb}
\usepackage{multirow}
\usepackage{mdframed}
\usepackage{booktabs}
\usepackage{float}

\definecolor{boxbg}{HTML}{F0F8FF} 
\definecolor{boxborder}{HTML}{4682B4} 
\definecolor{boxtitlebg}{HTML}{4682B4} 
\definecolor{boxtitletext}{HTML}{FFFFFF} 

\tcbset{
  myboxstyle/.style={
    colback=boxbg,
    colframe=boxborder,
    coltitle=boxtitletext,
    fonttitle=\bfseries,
    rounded corners,
    boxrule=1mm,
    left=1em,
    right=1em,
    top=1em,
    bottom=1em,
    enhanced,
    boxed title style={
      colback=boxtitlebg,
      colframe=boxborder,
      sharp corners,
      boxrule=0.5mm,
    },
    borderline={0.5mm}{-2mm}{boxborder},
  }
}

\usepackage[T1]{fontenc}
\usepackage[utf8]{inputenc}
\usepackage{microtype}
\usepackage{inconsolata}
\usepackage{graphicx}
\usepackage{hyperref}

\title{Adapting Large Language Models for \\Multi-Domain Retrieval-Augmented-Generation}

\correspondingauthor{alexandre.misrahi@epfl.ch, vassilina.nikoulina@naverlabs.com}

\authors{Alexandre Misrahi\textsuperscript{1*} \quad
Nadezhda Chirkova\textsuperscript{2} \quad
Maxime Louis\textsuperscript{2} \quad
Vassilina Nikoulina\textsuperscript{2} 
}
\affiliations{\textsuperscript{1}EPFL
 \textsuperscript{2}NAVER LABS Europe}
\contributions{}
\website{}
\websiteref{}

\teaserfile{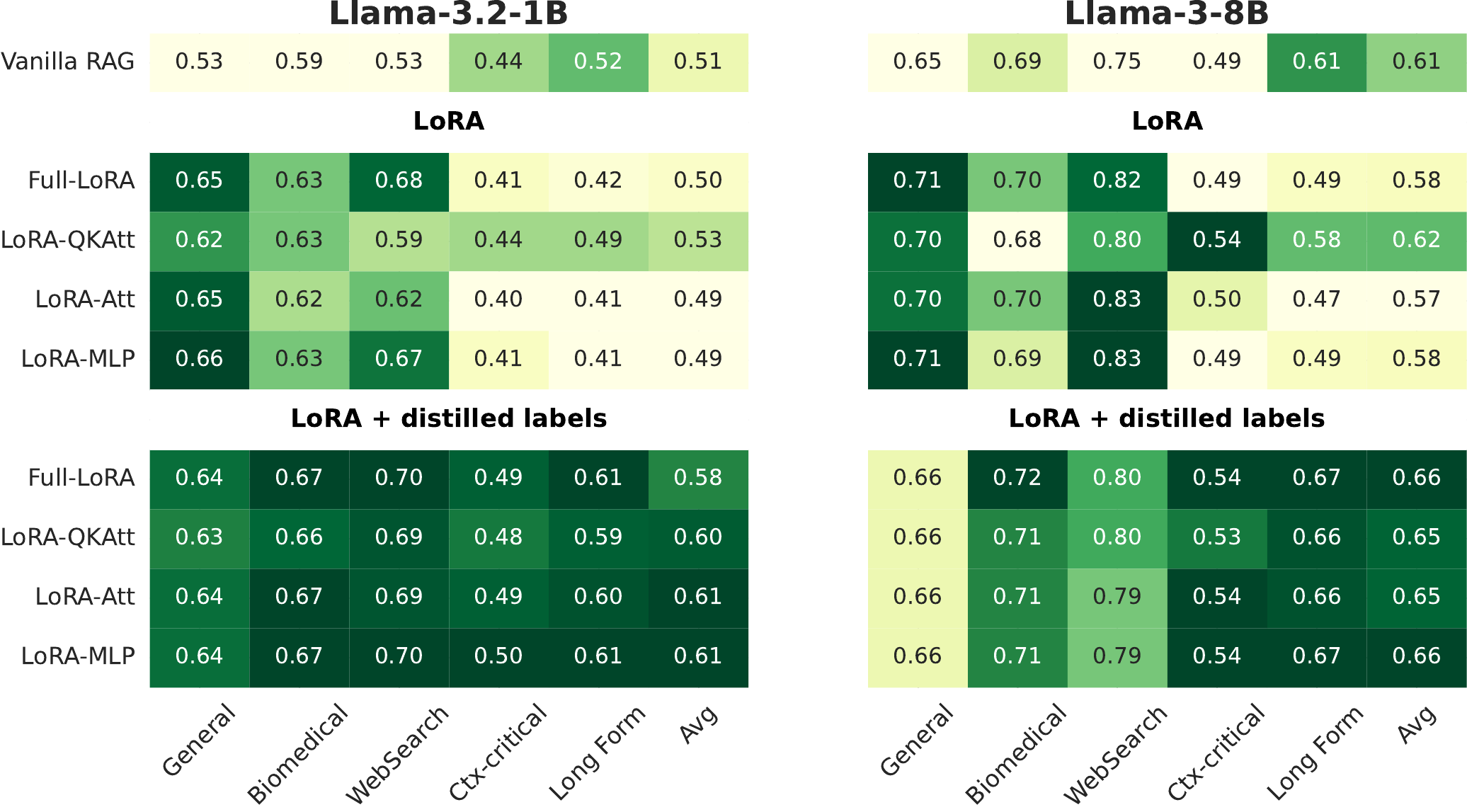}

\teasercaption{\label{fig:llama_rag_adaptation} RAG adaptation results, LLMEval. Color is relative to the corresponding column. Average column is computed across all individual datasets.}

\begin{abstract}
Retrieval-Augmented Generation (RAG) enhances LLM factuality, but multi-domain applications face challenges like lack of diverse benchmarks and poor out-of-domain generalization. The first contribution of this work is to introduce a diverse benchmark comprising a variety of question-answering tasks from 8 sources and covering 13 domains. Our second contribution consists in systematically testing out-of-domain generalization for typical RAG tuning strategies. While our findings reveal that standard fine-tuning fails to generalize effectively, we show that sequence-level distillation with teacher-generated labels improves out-of-domain performance by providing more coherent supervision. Our findings highlight key strategies for improving multi-domain RAG robustness.

\end{abstract}

\begin{document}

\maketitle
\begingroup
\renewcommand\thefootnote{*}
\footnotetext{Work done during internship at NAVER LABS Europe}
\endgroup

\section{Introduction}

Retrieval-Augmented Generation (RAG) \cite{DBLP:journals/corr/abs-2005-11401, gao2023retrieval} is a technique that enhances Large Language Models (LLMs) by retrieving relevant information from an external document repository to augment the input prompt. Often likened to an “open book” exam, this approach allows LLMs to ground their responses in external knowledge, thereby improving their faithfulness. RAG is especially valuable for out-of-domain queries—those beyond an LLM’s general knowledge—where the model must rely on its ability to exploit retrieved context.

While RAG is sometimes viewed as an alternative to supervised finetuning on domain-specific data, and has shown strong performance across various domains \cite{ovadia-etal-2024-fine}, recent work \cite{lin2024raditretrievalaugmenteddualinstruction, zhang2024raftadaptinglanguagemodel, ovadia-etal-2024-fine} has highlighted the benefits of finetuning LLMs specifically for RAG. This not only improves the model’s ability to use retrieved context effectively, but also increases robustness to irrelevant information. Our goal in this paper is to investigate RAG finetuning for out-of-domain tasks at test time.

Existing studies on multi-domain RAG \cite{jiao2024duetrag, zhang2024raftadaptinglanguagemodel, han2024ragqaarenaevaluatingdomain, han-etal-2023-robustqa, wei2024long} overlook two key aspects. First, existing multi-domain RAG benchmarks lack diversity and offer only limited insights into the multi-domain challenges. Second, to the best of our knowledge, none of the research performed in adapting LLMs for RAG consider domain-shift settings, when test-time queries or document repositories can differ from the ones used at training time in the style or domain.

\textbf{Our primary contribution is a carefully curated and highly diverse set of query datasets and corresponding document collections (datastores) covering multiple domains, task types and answer formats}\footnote{These datasets and document collections have been integrated into the \href{https://github.com/naver/bergen}{Bergen} benchmarking library for RAG.}. This resource enables a precise assessment of the limitations of the RAG pipeline and facilitates an analysis of the impact of retrievers, rerankers, and LLMs on overall performance. Our experimental study shows that performance improvement brought by RAG compared to zero-shot LLM prompting varies across domains, due to multiple factors such as variable complexity and genre of questions, datastore preprocessing, or retrieval performance. 

\textbf{Our second contribution is studying cross-domain generalization of various LLM finetuning strategies for RAG.} We show that commonly used tuning of an LLM for RAG on standard question answering datasets (with short labels) does not enable models to effectively use context in the presence of domain shift.  Our main finding is that general-domain sequence-level distillation using teacher-generated labels improves out-of-domain performance, likely due to the greater coherence of teacher-generated answers compared to ground truth labels.

Finally,\textbf{ we perform an in-depth analysis of results}, looking at various RAGChecker metrics \cite{ru2024ragcheckerfinegrainedframeworkdiagnosing} across multiple datasets to better understand the effect of different RAG adaptation techniques and support our hypothesis. 

\newcolumntype{M}{>{\begin{varwidth}{4cm}}l<{\end{varwidth}}} 

\begin{table*}[hbt!]
    \centering
    \resizebox{\textwidth}{!}{
    \begin{tabular}{|l|l|p{1in}|l|l|l|p{4in}|}
        \hline
        \textbf{Cat.} & \textbf{Dataset} & \textbf{Description} & \textbf{Query Source} & \textbf{Datastore Source} & \textbf{N queries} & \textbf{Example query \& document} \\
        \hline
        \multirow{5}{*}[-10ex]{\rotatebox[origin=c]{90}{BIOMEDICAL}} & Bioasq11b & Biomedical questions of types factoid/yesno/list & \makecell[l]{\href{https://huggingface.co/datasets/jenhsia/ragged}{Task B, 2023} \\\cite{Nentidis_2023}} & \multirow{2}{*}[-8ex]{\href{https://huggingface.co/datasets/jenhsia/ragged}{PubMed abstracts}} & 3837 & \makecell[l]{\textbf{Doc:} Biochemical purification of pseudopodia from migratory \\cells.: Cell migration requires [...] \\\textbf{Query:} Is it possible to purify pseudopodia to be used for \\proteomic analysis?\\\textbf{Label:} 'yes'}\\
        \cline{2-4}\cline{6-7}
        & Bioasq12b & Biomedical questions of types factoid/yesno/list & \makecell[l]{\href{http://participants-area.bioasq.org/datasets/}{Task B, 2024} \\\cite{447}} & & 940 & \makecell[l]{\textbf{Doc:} Deubiquitinating function of ataxin-3: insights from the \\solution structure of the Josephin domain.: Spinocerebellar ataxia \\type 3 is a human neurodegenerative disease [...] \\\textbf{Query:} Which is the protein implicated in Spinocerebellar ataxia \\type 3?\\\textbf{Label:} Ataxin-3} \\
        \cline{2-7}
        & CovidQA & Short and long-form QA related to covid-19 & \makecell[l]{\href{https://huggingface.co/datasets/deepset/covid_qa_deepset}{CovidQA} \\\cite{moller-etal-2020-covid}} & CORD-19 \cite{wang-etal-2020-cord} & 2019 & \makecell[l]{\textbf{Doc:} Hantaviruses in the Americas and Their Role as Emerging \\Pathogens [...]\\\textbf{Query:} Typically how long do the hamsters die post-inoculation?\\\textbf{Label:} 11 and 14-d}\\ 
        \hline
        \multirow{7}{*}[-28ex]{\rotatebox[origin=c]{90}{LONG-FORM}} & FiQA & Fact- and Opinion-based QA for finance & \makecell[l]{Task 2 of \\\href{https://sites.google.com/view/fiqa/}{FiQA challenge} \\\href{https://huggingface.co/datasets/LLukas22/fiqa}{Source}} & \makecell[l]{\href{https://huggingface.co/datasets/BeIR/fiqa}{BeIR (corpus)} \\\cite{thakur2021beir}} & 2561 & \makecell[l]{\textbf{Doc:} There are benefits associated with a cash only business (the \\link states a few). However [...]\\\textbf{Query:} A merchant requests that checks be made out to "Cash". \\Should I be suspicious?\\\textbf{Label:} There are benefits associated with a cash only business (the \\link states a few). However [...]}\\ 
        \cline{2-7}
        & Lifestyle & E.g. cooking, nutrition, everyday tasks. & \multirow{5}{*}[-20ex]{\makecell[l]{\href{https://github.com/awslabs/rag-qa-arena/tree/main/data}{Source} \\\cite{han-etal-2023-robustqa}}} & \multirow{2}{*}[-25ex]{\href{https://downloads.cs.stanford.edu/nlp/data/colbert/colbertv2/lotte.tar.gz}{LoTTE (StackExchange)}} & 2198 & \makecell[l]{\textbf{Doc:} A clove is not a clove. There are 2 main types of garlic: \\Hard Neck: hard neck garlic varieties [...]\\\textbf{Query:} what is a clove of garlic?\\\textbf{Label:} Each wedge of a garlic is known as a clove and the whole \\garlic is referred to as a head.}\\ 
        \cline{2-3}\cline{6-7}
        & Recreation & E.g. various video games. & &  & 2090 & \makecell[l]{\textbf{Doc:} Technically - you do not need to purchase anymore games. \\However, you will need to purchase and download [...]\\\textbf{Query:} if i buy garrys mod, will i need other games to play it?\\\textbf{Label:} You don't have to have any other games in order to \\actually play GMod. [...]}\\ 
        \cline{2-3}\cline{6-7}
        & Science & E.g. math, physics, biology. & &  & 1404 & \makecell[l]{\textbf{Doc:} (as a limit of partial sums). Now, when people then say stuff \\like $1 + 2 + 3 + \dots = -1/12$ they will  [...]\\\textbf{Query:} why does $1+2+3+\cdots = -\frac{1}{12}$? \\\textbf{Label:} There are numerous methods to determine that a particular \\result is correct. [...]}\\ 
        \cline{2-3}\cline{6-7}
        & Technology & E.g. security, hardware, software. & &  & 2064 & \makecell[l]{\textbf{Doc:} The easiest way to have the Finder refresh its listing is to \\enter a subfolder [...]\\\textbf{Query:} is there a way to refresh a finder file listing? \\\textbf{Label:} There are a number of approaches, one is to use a simple \\AppleScript [...]}\\ 
        \cline{2-3}\cline{6-7}
        & Writing & E.g. syntax, grammar, vocabulary. & &  & 2694 & \makecell[l]{\textbf{Doc:} Fact and fairy aren't etymologically related, but John \\Lawler's answer is [...]\\\textbf{Query:} etymology of fairy \\\textbf{Label:} The "-ry" part of the word fairy is derived from the \\"-ery" suffix [...]}\\ 
        \hline
        \multirow{1}{*}[5ex]{\rotatebox[origin=c]{90}{WEB-SEARCH}} & SearchQA & General QA with context from search engine & \makecell[l]{\href{https://huggingface.co/datasets/kyunghyuncho/search_qa}{Source} \\\cite{DBLP:journals/corr/DunnSHGCC17}} & \href{https://huggingface.co/datasets/kyunghyuncho/search_qa}{Search Engine Results} & 21613 & \makecell[l]{\textbf{Doc:} Jun 24, 2010 ... Good King Wenceslas was murdered by his \\brother. [...]\\\textbf{Query:} This "Good King" of Bohemia was killed by his \\brother Boleslav while on the way to mass \\\textbf{Label:} Wenceslas}\\ 
        \hline
        \multirow{1}{*}[-8ex]{\rotatebox[origin=c]{90}{CONTEXT—CRITICAL}} & ParaphraseRC & Short, fact-based movie QA (pre-pended with movie name). & \makecell[l]{\href{https://huggingface.co/datasets/ibm/duorc/viewer/ParaphraseRC}{ParaphraseRC} \\\cite{DBLP:journals/corr/abs-1804-07927}} & \href{(https://huggingface.co/datasets/ibm/duorc/viewer/ParaphraseRC}{Movie plots} & 13111 & \makecell[l]{\textbf{Doc:} On the Path: Luna and Amar are a young Bosnian couple \\living in Sarajevo. [...]\\\textbf{Query:} On the Path: Why did Amar loses his job for being at \\work?\\\textbf{Label:} Amar loses his job for being drunk at work}\\ 
        \cline{2-7}
        & SyllabusQA & Short questions about course logistics & \makecell[l]{\href{https://github.com/umass-ml4ed/SyllabusQA/tree/main/data/dataset_split}{Source} \\\cite{fernandez2024syllabusqacourselogisticsquestion}} & \href{https://github.com/umass-ml4ed/SyllabusQA/tree/main/syllabi/syllabi_redacted/text}{Courses syllabi} & 957 & \makecell[l]{\textbf{Doc:} MUSIC-ED 500KU Syllabus S23: l be weighted as follows: \\Assignment Percentage of Final Grade Midpoint Reports [...]\\\textbf{Query:} UMASS-Math3312023: How many credits will I earn for \\this course? \\\textbf{Label:} No/insufficient information}\\ 
        \cline{2-7}
        & TechQA & Technical support queries & \makecell[l]{\href{https://huggingface.co/datasets/rojagtap/tech-qa}{Source} \\\cite{castelli-etal-2020-techqa}} & \href{https://huggingface.co/datasets/rojagtap/tech-qa}{IBM Technotes} & 621 & \makecell[l]{\textbf{Doc:} IBM Solving IBM MQ Java code version mismatches using \\the mqjavalist.sh script  [...]\\\textbf{Query:} How do I tell when there are mismatched MQ jars in my \\application server? [...]\\\textbf{Label:} Finding and eliminating duplicate copies of the MQ jar \\files can be a difficult task [...]}\\ 
        \hline
    \end{tabular}
    }
    \caption{Multidomain RAG benchmarks}
    \vspace{-0.5em}
    \caption*{\textit{The datastores for ParaphraseRC, CORD-19, LoTTE are processed in chunks of 100 words with 20 words overlap between consecutive chunks. The datastores for SyllabusQA and TechQA are processed in chunks of 1000 characters with 200 characters overlap between consecutive chunks. For SyllabusQA and ParaphraseRC, each chunk is pre-pended with the document title (course title and movie title, respectively). }}
    \label{tab:datasets}
\end{table*}

\section{Related work}
\label{section:related_work}

\paragraph{Multi-domain RAG benchmarks.} While some papers propose benchmarks for evaluating retrieval-augmented LLMs in multi-domain settings  \cite{han2024ragqaarenaevaluatingdomain, han-etal-2023-robustqa, wei2024long, zhang2024raftadaptinglanguagemodel}, we argue that existing multi-domain RAG benchmarks lack domain diversity and do not capture a wide variability of real-world scenarios where RAG can be useful. In particular, existing benchmarks usually include a small number of domains, crafted from a limited set of sources (usually 1--3 sources), and often focus on a particular kind of answers, e.g. only include long-form answers. In our work, we introduce a diverse multi-domain RAG benchmark, which includes data from 8 sources, covers 13 domains in total, and contains questions and answers of various types and genres.

\paragraph{Out-of-Domain Generalization in NLP.} Generalization of NLP pipelines is an active area of study \cite{ijcai2021p628, wang2022measureimproverobustnessnlp, ailem2024examiningrobustnessllmevaluation}. \citet{yang-etal-2023-glue} examines out-of-distribution (OOD) generalization of pretrained LLMs across domains for different tasks, and observe a significant gap between LLM and human OOD performance. In \cite{calderon2024measuringrobustnessnlpmodels}, the authors evaluate domain robustness of models and in particular show that zero-shot LLMs  demonstrate superior cross-domain robustness compared to fine-tuned models -- we will show similar conclusions for RAG tasks. 

\paragraph{Techniques for Improving RAG Pipelines.}
Recent efforts to improve RAG pipeline performance have focused on optimizing both retrieval and generation components. RA-DIT \cite{lin2024raditretrievalaugmenteddualinstruction} jointly tunes the retriever and generator, enhancing overall performance. Instruct-Retro \cite{wang2024instructretroinstructiontuningpost} introduces a two-stage approach by continuing pre-training with RAG, followed by an instruction-tuning phase, which improves post-instruction tuning performance. RAFT \cite{zhang2024raftadaptinglanguagemodel} has shown that finetuning an LLM for RAG on oracle documents mixed with noisy contexts improves robustness at test time. Authors claim better usage of context is enabled by Chain-of-Thought (CoT) augmentation of the training data. However, such CoT augmentation requires processing whole set of samples in training data, and also expects oracle documents to be available at training time which might be true only for a very limited set of datasets. While these methods have demonstrated success, they are usually tested only in-domain, i.e. the training and the testing data come from the same distribution. In contrast, in our work, we study cross-domain generalization of various LLM adaptation techniques for RAG.

\section{Multidomain RAG Benchmark}
\label{sec:benchmark}
\begin{figure}
    \centering
    \includegraphics[width=0.85\linewidth]{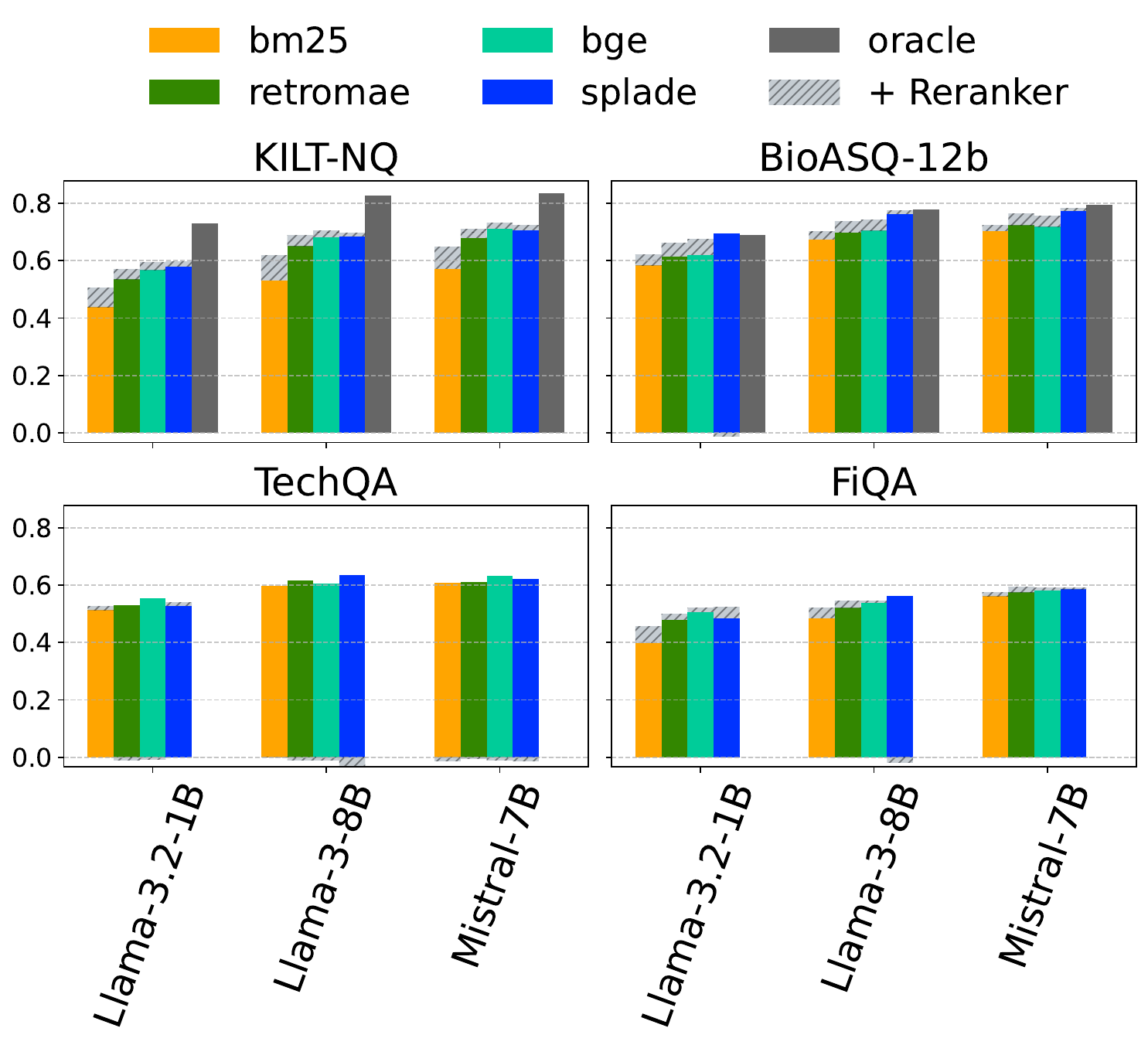}
    \caption{LLMEval scores for three LLM generators evaluated across four datasets and multiple retriever/reranker configurations. Oracle documents shown where available.. Full results given in Appendix Table \ref{tab:fig1_full_table}.}
    \label{fig:fig1}
\end{figure}

We address the limitations of existing out-of-domain benchmarks mentioned in \S \ref{section:related_work} by building a new diverse benchmark suite for RAG. Specifically, we include multiple domains, multiple formats of answers and questions, multiple collections of documents from various sources, and sets where the relative importance of the context is variable.  The selected Question Answering (QA) datasets and associated document collections are summarized in Table \ref{tab:datasets} and described below. 

\paragraph{Biomedical QA. }
The BioASQ challenge \cite{Nentidis_2023} datasets from 2023 and 2024 consist in biomedical QA. We further contextualize BioASQ queries with the most relevant of $\sim$ 58M PubMed abstracts. The biomedical domain also includes CovidQA \cite{moller-etal-2020-covid}, a dataset of long-form QA about the Covid-19 pandemic. The datastore for CovidQA is the CORD-19 dataset. Compared to BioASQ, this dataset requires more specific biomedical knowledge, and in general we observe lower absolute metrics on CovidQA compared to BioASQ. In the biomedical setting, the robustness challenge stems from rare, out-of-distribution tokens. 

\paragraph{Web-Search QA. }

SearchQA is a question-answering dataset augmented with text snippets retrieved from a search engine. 
The difficulty in this dataset stems from the convoluted phrasing of the queries posed as ``Trivia'' questions. 
For example: ``Slow down, as in a car'' yields gold label ``decelerate''. 

\paragraph{Context-critical QA. }
ParaphraseRC consists of short questions about movies, with datastores consisting of movie plot chunks. SyllabusQA is made of short questions about university course syllabi. TechQA corresponds to the highly specialized domain of company- and tool-specific forums. All of these domains contain queries that can only be replied to with the correct information from the relevant documents and cannot be replied to from the internal LLM knowledge alone. 
These domains are particularly helpful for evaluating the retriever and generator's ability to extract relevant information from noisy context. For example, with the query ``CS-101: How is this course graded?'' in SyllabusQA, the retriever may retrieve chunks relevant either to the course CS-101, or chunks relevant to how certain courses are graded. In the latter case, the task might be challenging for the generator to correctly identify the chunk specific to CS-101, especially in the zero-shot setting (no fine-tuning on SyllabusQA). 

\paragraph{Long-Form QA. }

Finally, we adopt several long-form question answering datasets which are more open-ended QA tasks. This includes the following domains from the RobustQA collection \cite{han-etal-2023-robustqa}: Finance, Lifestyle, Recreation, Science, Technology, Writing.

\subsection{Experimental settings}

We use BERGEN \footnote{\url{https://github.com/naver/bergen}} \cite{rau2024bergenbenchmarkinglibraryretrievalaugmented} to benchmark multi-domain RAG on QA tasks. 

 \paragraph{General setup. } Unless otherwise specified, we use \texttt{SPLADE-v3} \cite{splade-v3} retriever to identify a first set of relevant documents given a query. These documents are further reranked using \texttt{DeBERTa-v3} \cite{he2021debertav3}, a cross-encoder computing relevance score for each document relative to the query. For generation, we use the instruct versions of Llama-3.2-1B\footnote{\href{https://huggingface.co/meta-llama/Llama-3.2-1B-Instruct}{meta-llama/Llama-3.2-1B-Instruct}}, Llama-3-8B\footnote{\href{https://huggingface.co/meta-llama/Meta-Llama-3-8B-Instruct}{meta-llama/Meta-Llama-3-8B-Instruct}} and Mistral-7B\footnote{\href{https://huggingface.co/mistralai/Mistral-7B-Instruct-v0.2}{Mistral-7B-Instruct}}. 
    
 \paragraph{Evaluation.} To evaluate generated responses, we mostly use LLM evaluation, denoted as LLMEval\footnote{LLMEval prompts an open-source LLM to output a binary judgment about the correctness of the generated response, given the input question and the ground truth labels.}, but we also consider Match\footnote{Match measures if any of the ground truth labels is contained verbatim in the generated response as a \textit{substring}.} and Recall\footnote{Recall measures a percentage of words from the ground truth labels that are contained verbatim in the generated response.}. LLMEval was shown to be better correlate with GPT-4 evaluations than match-based metrics in \cite{rau2024bergenbenchmarkinglibraryretrievalaugmented}. Similar to Match metrics, LLMeval is convenient to interpret since it reports the portion of successful replies. LLMEval is particularly useful for comparing long generations/ground truth answers, since Match will always output the zero evaluation result,  and Recall is highly impacted by common words, and hard to interepret. Details about the LLMEval, as well as some examples of evaluation, are given in Appendix \ref{appendix:llm_eval}. 
 
\section{Zero-Shot RAG results}
\subsection{RAG robustness to the context.} First, we analyse how good different LLMs are at exploiting context in RAG settings. Figure \ref{fig:fig1} compares how different retrievers and rerankers impact quality of answer generation. We add oracle documents when available to assess the performance of the retrieval pipeline. We select four datasets representing different domains documented in the table \ref{tab:datasets}: (1) NQ for general domain \cite{kwiatkowski2019natural}, (2) BioASQ-12b for  biomedical domain, (3) TechQA for Context-critical and (4) FiQA for Long-form.  

\paragraph{Effect of retriever/reranker.}
For all domains, we observe systematic performance gains with stronger retrievers, i.e. model-based retrievers vs bm25. Furthermore, the second stage reranking improves performance for most of the cases. The behaviour is similar both for the small model (Llama3.2-1B) and larger models (Llama3-8B, Mistral-7B). At the same time, even with oracle documents, none of the models are able to fully exploit relevant context, as there is still 20\%-30\% gap in performance to reach the perfect answer, which highlights importance of LLM adaptation. 

\begin{figure}[t]
    \centering
    \includegraphics[width=0.85\linewidth]{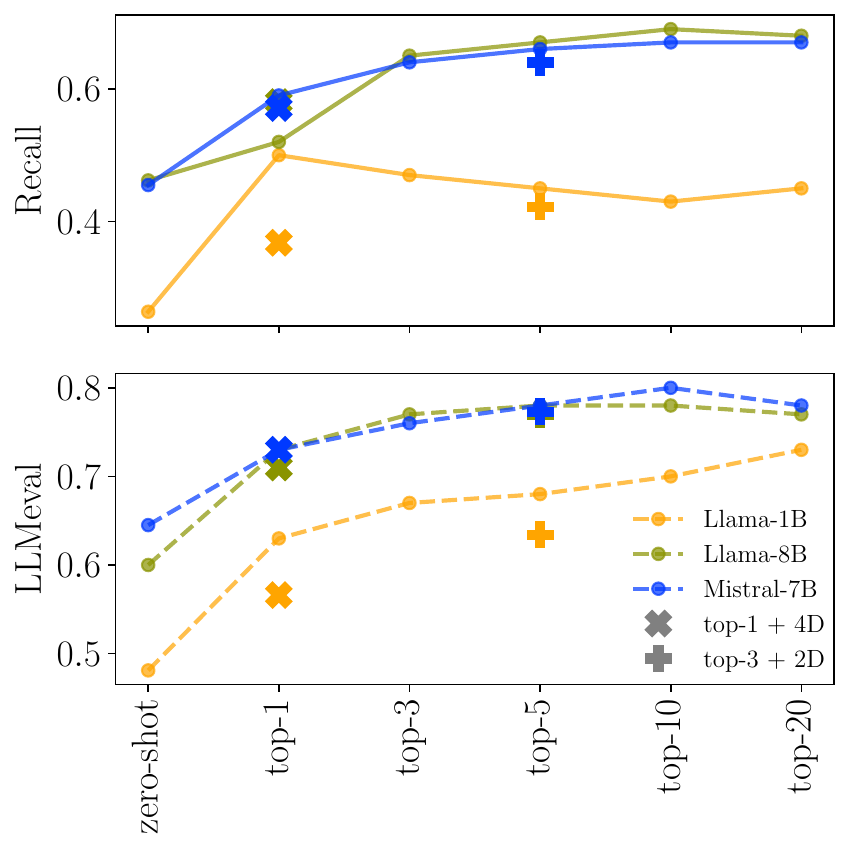}
    \caption{BioASQ-12b across Top-k Retrieved Documents (with splade-v3 retriever, DeBERTa-v3 reranker). $D$ denotes distractor documents chosen at random from PubMed abstracts to add noise to the context.  The smaller model is more sensible to noise.}
    \label{fig:bioasq-top-k}
\end{figure}

 \paragraph{Importance of top-k documents.} Figure \ref{fig:bioasq-top-k} compares how different LLMs handle the increased amount of retrieved documents provided in the context. All models benefit from larger number of documents when assessed with LLMEval. In addition to varied top-k documents, we also assess LLM ability to identify relevant content when distractor documents are added in the context. We note that large models (Llama3-8B and Mistral-7B) are relatively robust to the noise introduction, however the small model (Llama3.2-1B) is more sensitive to noise. 
 
\subsection{RAG domain robustness}

\begin{table*}[hbt!]
    \begin{subtable}{\textwidth}
    \centering
    \captionsetup{justification=centering}
    \begin{tabular}{l|l|cc|cc}
       \multirow{3}{*}{\textbf{Domain }}& \multirow{3}{*}{\textbf{Dataset}} & \multicolumn{2}{c|}{\textbf{Llama-3.2-1B-Instruct}} & \multicolumn{2}{c}{\textbf{Llama-3-8B-Instruct}} \\
       
         & &  \textbf{Without RAG} & \textbf{With RAG} & \textbf{Without RAG} & \textbf{With RAG} \\
        \hline
                \multirow{3}{*}{Biomedical}& Bioasq12b & 0.48 & 0.68 & 0.64 & 0.76 \\
        & CovidQA & 0.38 & 0.50 & 0.46 & 0.61 \\
        & \textbf{Mean} & \textbf{0.43} & \textbf{0.59} & \textbf{0.55} & \textbf{0.69}\\\hline
        \multirow{7}{*}{Long-form}& FiQA & 0.54 & 0.50 & 0.55 & 0.51 \\
        & Lifestyle & 0.50 & 0.56 & 0.51 & 0.62 \\
        & Recreation & 0.38 & 0.45 & 0.47 & 0.58 \\
        & Science & 0.49 & 0.49 & 0.33 & 0.64 \\
        & Technology & 0.47 & 0.53 & 0.43 & 0.64 \\
        & Writing & 0.44 & 0.49 & 0.53 & 0.67 \\
        & \textbf{Mean} & \textbf{0.47} & \textbf{0.50} & \textbf{0.47} & \textbf{0.61} \\ \hline
        Web-Search & SearchQA & \textbf{0.38} & \textbf{0.57} & \textbf{0.55} & \textbf{0.75} \\\hline
         \multirow{4}{*}{Context-critical}& ParaphraseRC & 0.18 & 0.47 & 0.32 & 0.63 \\
        & SyllabusQA & 0.37 & 0.30 & 0.40 & 0.26 \\
        & TechQA & 0.52 & 0.54 & 0.49 & 0.59 \\
        & \textbf{Mean} & \textbf{0.36} & \textbf{0.44} & \textbf{0.40} & \textbf{0.49} \\ 
        \hline
    \end{tabular}
    \end{subtable}
    \caption{Benchmarking RAG with models of different sizes (LLMEval). The RAG pipeline uses \texttt{SPLADE-v3} retriever and \texttt{DeBERTa-v3} reranker.}
    \label{tab:benchmark-results-llama1B-llama8B_llmeval}
\end{table*}

Table \ref{tab:benchmark-results-llama1B-llama8B_llmeval} reports RAG performance (LLMeval) across various domains on off-the-shelf LLMs which are not fine-tuned for RAG tasks (Recall metric shown in Appendix Table \ref{tab:benchmark-results-llama1B-llama8B}). First, we observe that the use of RAG improves performance compared to no-RAG on almost all datasets and for both Llama-1B and Llama-8B. Second, gains with RAG are about equivalent for Llama-1B and Llama-8B, except on the Long-form generation tasks, where Llama-8B benefits significantly more (+14\% versus +3\% in LLMEval for Llama-1B) from the addition of the retrieved context. On the long-form tasks, a good generation must be able to use most parts of the provided context, and not simply identify a specific fact as for short answers. This is a likely reason as to why the larger model benefits more from the provided RAG context. Third, improvements from RAG are heterogeneous across the different datasets and tasks: it brings +14\% for biomedical tasks, +20\% for Long-form tasks but only 9\% for Context-Critical tasks for Llama-8B, and similar relative gains for Llama-1B. This indicates different generalization abilities of the RAG pipeline to the different domains. Additional results, confirming these conclusions, on other LLMs can be found in appendix Table \ref{tab:benchmark-results-gemma-qwen} \& \ref{tab:benchmark-results-llama-solar}.

\section{RAG adaptation} 
\label{sec:ragadapt}

\subsection{LLM Finetuning for RAG }

Recent works have demonstrated that it is beneficial to finetune an LLM to encourage it to better use retrieved context \cite{rau2024bergenbenchmarkinglibraryretrievalaugmented, lin2024raditretrievalaugmenteddualinstruction, wang2024instructretroinstructiontuningpost, liu2024chatqa}. It has been shown to improve RAG performances in-domain, and we investigate whether this holds for out-of-domain RAG generalization. To do so, we run supervised fine-tuning on the MultiQA dataset\footnote{\href{https://huggingface.co/datasets/dmrau/multi_qa}{huggingface/datasets/dmrau/multi\_qa}}: a dataset consisting of 450k general domain questions and answers, described in Appendix Table \ref{tab:multiqa-content}. Its associated document collection consists of Wikipedia \cite{petroni2020kilt} and MSMARCO \cite{bajaj2016ms} documents. Each supervised fine-tuning sample then consists of a prompt (taken from \cite{rau2024bergenbenchmarkinglibraryretrievalaugmented}) with 5 retrieved documents, the question and its answer. Models are trained with LoRA \cite{hu2022lora}. After training, we evaluate the RAG-adapted models on the multi-domain benchmark. Training hyper-parameters are given in Appendix \ref{appendix:ft_hyperparameters}.

\begin{table*}
    \centering
    \begin{tabular}{c|ccccc|c}\hline
         &  General&  Biomedical&  WebSearch&  Cxt-critical&  Long Form& Avg\\\hline
 \multicolumn{7}{c}{Llama3.2-1B}\\\hline
         Vanilla RAG&  0.53&  0.59&  0.53&  \textbf{0.44}&  \textbf{0.52}& \textbf{0.51}\\
         FT with RAG&  \textbf{0.65}& \textbf{ 0.63}&  \textbf{0.68}&  0.41&  0.42& 0.50\\\hline
         \multicolumn{7}{c}{Llama3-8B}\\\hline
         Vanilla RAG&  0.65&  0.69&  0.75&  0.49&  \textbf{0.61}& \textbf{0.61}\\
         FT with RAG&  \textbf{0.71}& \textbf{ 0.70}&  \textbf{0.82}&  0.49&  0.49& 0.58\\\hline
         \multicolumn{7}{c}{Mistral-7B}\\\hline
 Vanilla RAG& 0.69&\textbf{ 0.70}& 0.72& \textbf{0.52}& \textbf{0.65}&\textbf{0.64}\\
 FT with RAG&\textbf{ 0.72}& 0.69& \textbf{0.85}& 0.50& 0.53&0.59\\\hline
    \end{tabular}
    \caption{Comparing results (LLMEval) of vanilla RAG (out-of-the-box LLM) vs finetuning with RAG (LoRA FT on MultiQA dataset). We report average score across multiple datasets within the same domain. Detailed results per dataset are available in Appendix Tables \ref{tab:rag_adaptation_full_general} - \ref{tab:rag_adaptation_full_avg}.}
    \label{tab:default_ft}
\end{table*}

Table \ref{tab:default_ft} compares vanilla RAG to RAG-adapted model. We note that while RAG adaptation does indeed improve on General and WebSearch domains, it degrades performance on Context-critical and Long-form datasets. The performance on biomedical datasets gets improved for Llama3.2-1B model, but doesn't really change for larger models.  \textbf{This indicates that standard RAG adaptation does not generalize beyond the training data.}

\subsection{Robust RAG adaptation}
If we want RAG finetuning to transfer across domains, it would mean that it should learn to identify relevant information in the retrieved documents, that would transfer to different types of collections, and different types of answers. 
Therefore the drop on certain domains can be explained by multiple factors:
\begin{itemize}
    \item The different distribution of relevant information in the retrieved documents can affect the generator's ability to exploit  context, 
    \item The stylistic differences in questions can affect the model's ability to understand the task at hand, 
    \item Overfitting to the answer style in the training data may hurt performance on long-form datasets. 
\end{itemize}
In this work we compare different adaptation strategies to improve robustness under domain shift. 

\paragraph{Improving attention pattern via LoRA-QK finetuning.}
 We hypothesize that in order to learn model to better focus on relevant documents we need to modify the attention pattern.  Therefore we train a LoRA variant that only updates $Q$ and $K$ matrices (\textit{LoRAQKAtt}) of attention which are responsible for the attention pattern. We also train LoRA variants updating MLP layers only (\textit{LoRAMLP}), and LoRA updating all the attention matrices (\textit{LoRAAtt}) for comparison. We adjust the rank of different LoRA variants in order to keep a similar amount of trainable parameters for fair comparison (details in Appendix Table \ref{tab:lora_hyperparameters}). 

\paragraph{Sequence-level Knowledge Distillation.}
Most of the gold answers of standard datasets that compose MultiQA dataset correspond to very short text snippets, without any additional information or explanation. Standard answers generated by an instruction-tuned LLMs tend to be in a quite different format: more verbose, containing more details and explanations. Therefore finetuning on short labels forces the model to substantially move away from its initial generation distribution. In order to avoid this effect we rely on Sequence-level Knowledge Distillation (SKD) \cite{kim-rush-2016-sequence}. It consists in training the student model on the labels generated by a strong teacher, rather than on the gold labels. \cite{Zhou2020Understanding} suggests that SKD allows to reduce the complexity of the data and therefore makes learning simpler for the student. In our experiments, we rely on \textbf{Mistral-7B} model to generate labels. Preliminary experiments showed Mistral-7B is an excellent teacher, on par with findings in \cite{xu2024stronger, louis2025pisco}. We emphasize that student model training is performed with the same context as used in teacher generations.

We also report additional regularization experiments with distractors in the Appendix \ref{app:distractors}.

\subsection{Results and Analysis.}

Figure \ref{fig:llama_rag_adaptation} compares different proposed strategies for robust RAG adaptation. 

\paragraph{Improved attention pattern.} \textbf{We note that \textit{LoRA-QKAtt} is indeed more resilient to distribution shift} on \textit{Context-critical }and \textit{Long-form} datasets compared to other LoRA variants: for Llama-3-8B it does improve over the baseline on \textit{Context-critical} datasets which follow different distribution of relevant information in the retrieved documents.  It also better preserves performance on Long-Form datasets compared to other FT strategies. Table \ref{tab:llmeval_oracle} further confirms that \textit{LoRA-QKAtt} is better at exploring oracle context compared to Vanilla RAG and more robust compared to other FT variants.

\paragraph{Sequence Knowledge Distillation.}
\textbf{Knowledge distillation slightly decreases performance on general domain, but does improve across all other domains and lead to the best results overall.} We note, that when adaptation is performed on distilled labels, the adaptation strategy doesn't really matter, and all LoRA variants perform similarly. We view two reasons for this. First, we believe that this is due to the fact that there is less discrepancy between the data used for adaptation and models' natural distribution. Therefore, model adaptation is fully focused on better context exploitation rather than mimicking training data format.  Table \ref{tab:llmeval_oracle} provides additional evidence for this, demonstrating that model FT with distilled labels exploits better relevant context provided by Oracle documents. Second, teacher-generated labels include explanations and reasoning from the context, much like in \cite{zhang2024raftadaptinglanguagemodel}. Note that, at variance with \cite{zhang2024raftadaptinglanguagemodel} which generates chain-of-thoughts for fine-tuning, distillation labels don't rely on the existence of an oracle document. These labels are self-consistent because generated from the same retrieved context used at train time.

\begin{table}
\centering

\begin{tabular}{c|cc|cc}
 & \multicolumn{2}{c}{Llama-3.2-1B}& \multicolumn{2}{c}{Llama-3-8B}\\
& \small{KILT-NQ} & \small{BioASQ12b} & \small{KILT-NQ} &  \small{BioASQ12b}  \\\hline
\small{Vanilla} & 0.73 & 0.69 & 0.83 & 0.78 \\\hline
& \multicolumn{4}{c}{FT with LoRA}\\\hline
\small{Full} & \textbf{0.77} & 0.67 & 0.82 & 0.75 \\
\small{QKAtt} & 0.74 & \textbf{0.74} & \textbf{0.85} &\textbf{ 0.79} \\
\small{Att} & 0.76 & 0.71 &\textbf{ 0.85} & 0.76 \\
\small{MLP} & \textbf{0.77 }& 0.70& 0.82 & 0.75 \\\hline
 & \multicolumn{4}{c}{FT with LoRA + distilled labels}\\\hline
\small{Full} & \textbf{0.80}&\textbf{ 0.77} &\textbf{ 0.86} &\textbf{ 0.81} \\

\end{tabular}
\caption{LLMeval scores for RAG with oracle documents in the context,  with and without FineTuning for RAG. We report Recall scores in the table \ref{tab:recall_oracle} in the Appendix}
\label{tab:llmeval_oracle}
\end{table}

\paragraph{In-depth analysis with RagChecker.}

\begin{figure*}
\begin{tabular}{cc}
\includegraphics[width=7.5cm]{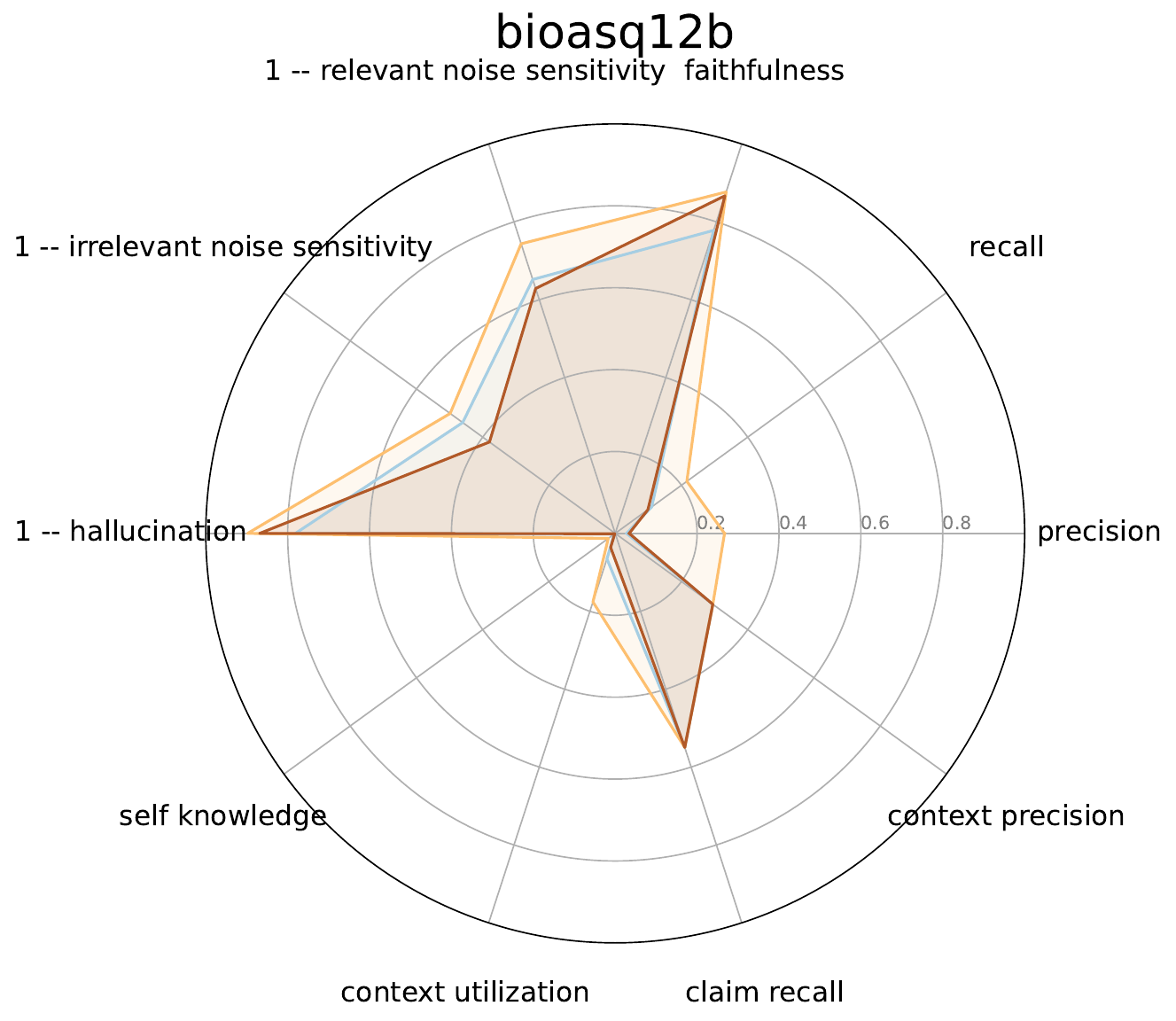} &
\includegraphics[width=7.5cm]{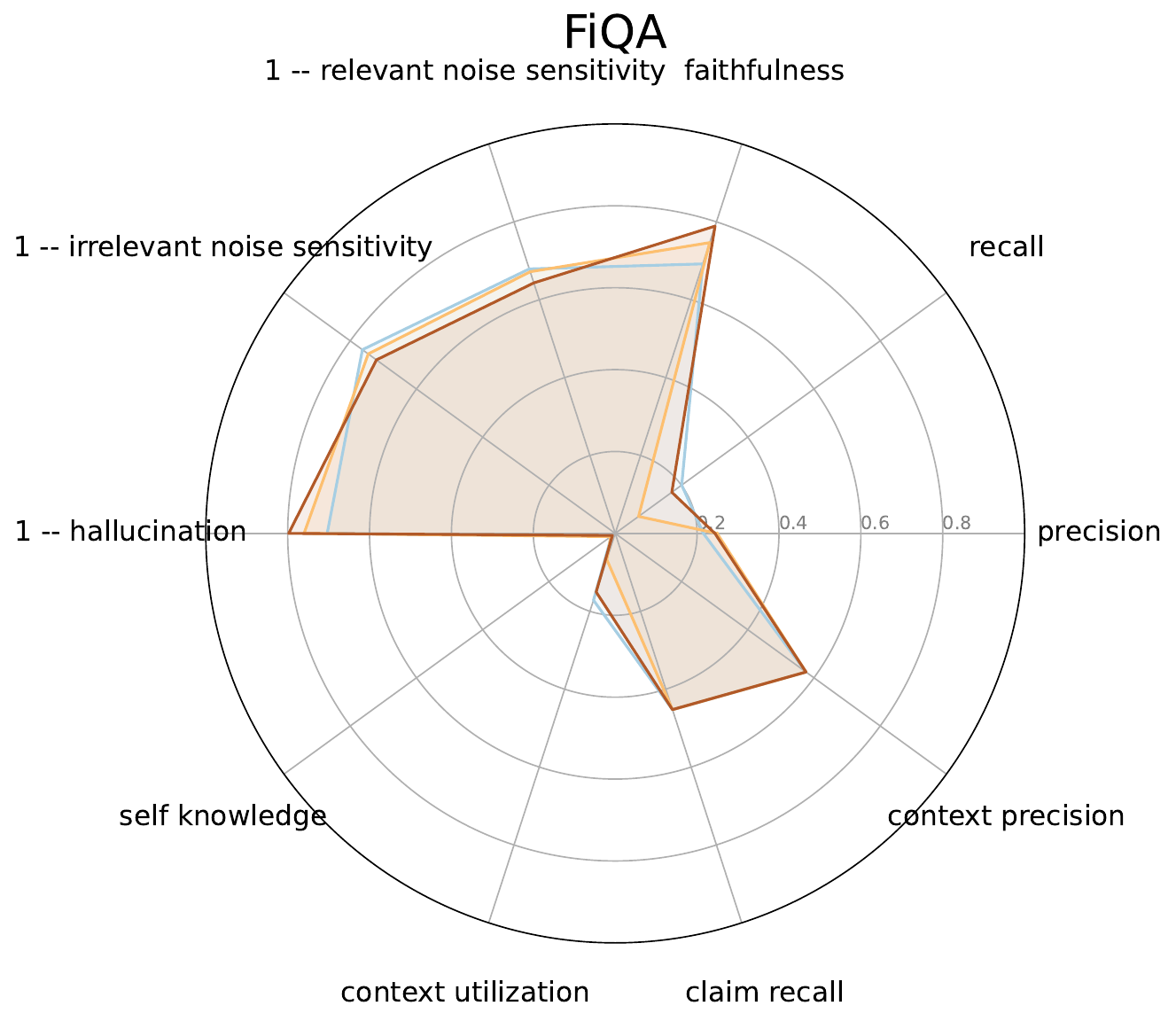} \\
\includegraphics[width=7.5cm]{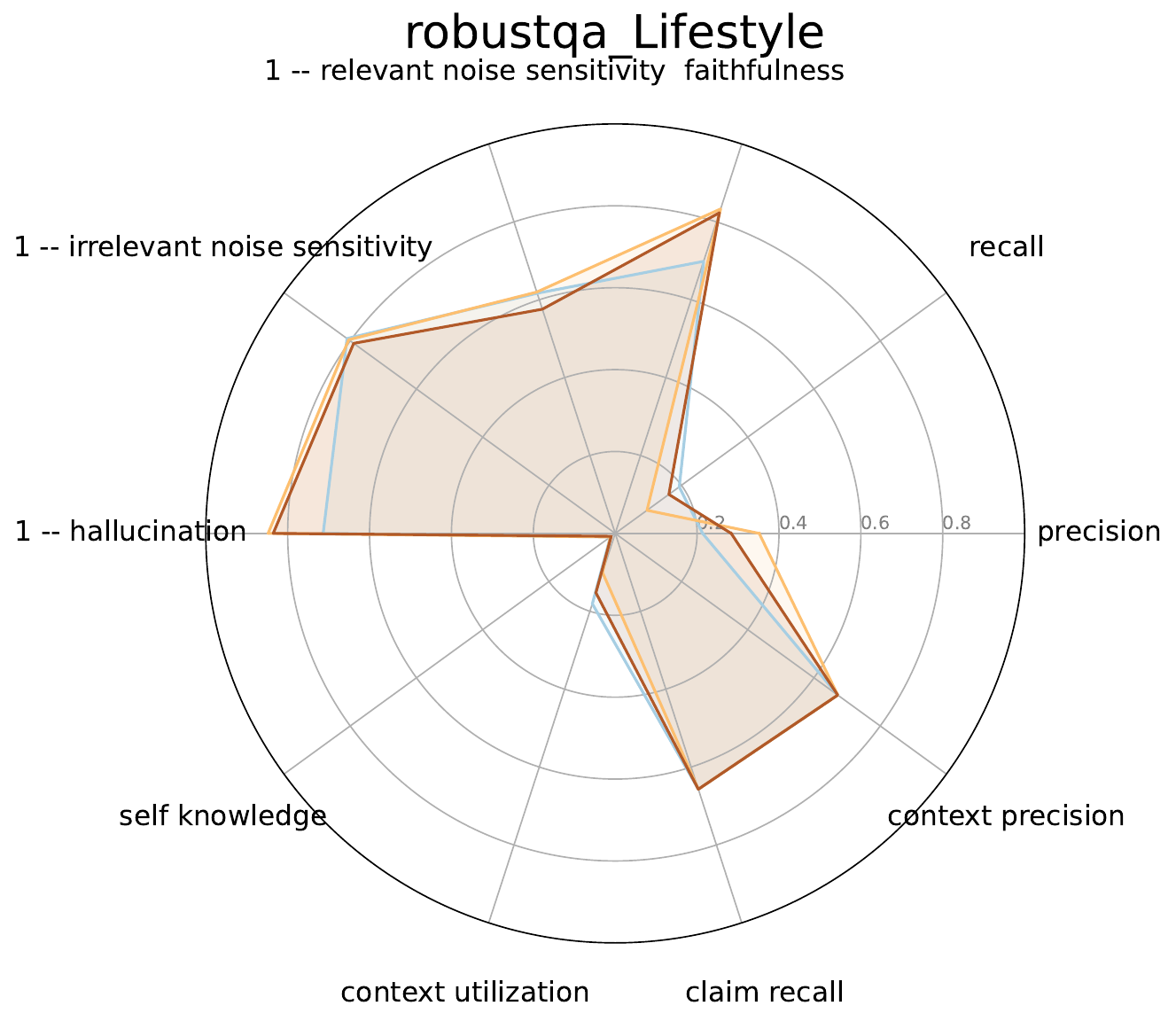} &
\includegraphics[width=7.5cm]{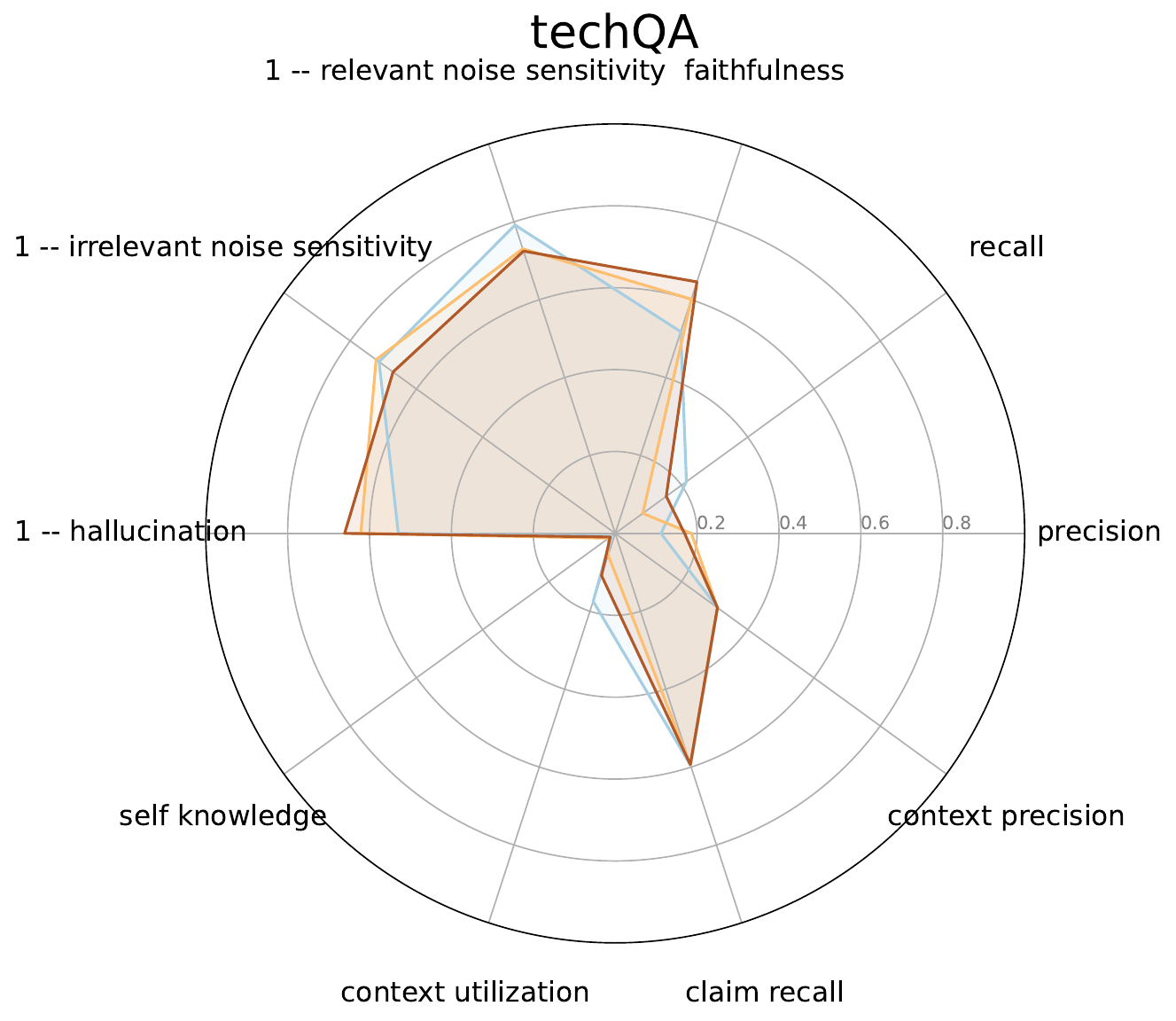} \\
\multicolumn{2}{c}{\includegraphics[width=5.5cm]{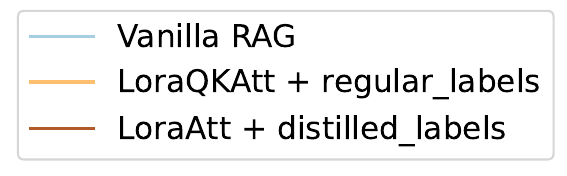}}
\end{tabular}
\centering
\caption{RAGChecker metrics on a subset of domains for Llama-3.2-1B. Effects on other domains are identical to one of the reported domains. All metrics are "the higher the better". The BioASQ dataset has short ground truth labels while other datasets have long labels. To compute metrics we use a Qwen claim extractor to segment out claims made by the generator, and a RoBERTa claim entailment checker.\protect\footnotemark}
\label{fig:ragchecker_metrics}
\end{figure*}
\footnotetext{See \href{https://huggingface.co/Qwen/Qwen2.5-3B-Instruct}{Qwen/Qwen2.5-3B-Instruct} and \href{https://huggingface.co/ynie/roberta-large-snli_mnli_fever_anli_R1_R2_R3-nli}{ynie/roberta-large-snli}}

\label{sec:ragchecker}
To analyze deeper various aspects of the RAG pipeline, we adopt a set of metrics introduced in RAGChecker \cite{ru2024ragcheckerfinegrainedframeworkdiagnosing}. This framework relies on two auxiliary models, namely a \textit{claim extractor} and an \textit{entailment classifier}. RAGChecker first extracts atomic claims from ground truth answers and generated responses, and then defines a set of metrics based on the entailment classification results between atomic claims and various pieces of text (retrieved context, ground truth answer, or generated response). We refer the reader to \cite{ru2024ragcheckerfinegrainedframeworkdiagnosing} for the full definition of each particular metric. Figure \ref{fig:ragchecker_metrics} reports RAGChecker metrics to compare three best models from Figure~\ref{fig:llama_rag_adaptation}, on a subset of domains. Results for other domains are identical.

Domains with long ground truth labels (e.g. FiQA, TechQA) exhibit similar trends and show  evidence that the \textit{model trained with distilled labels features a higher reliance on retrieved contexts than the baseline zero-shot model}.
This is emphasized by consistent improvement in metrics \textit{Faithfulness} (i.e. higher percentage of response claims entailed by the context) and \textit{Hallucination} (i.e. lower percentage of response claims not entailed by the retrieved context or by the ground truth), but also by the reduction in metrics measuring sensitivity to the noise in retrieved contexts (i.e. higher percentage of response claims not entailed by the ground truth and entailed by the context). Higher reliance on the context is exactly the effect we are aiming to achieve with our tuned models. However, we highlight that the observed differences in metric values can also be partly attributed to a different \textit{style} in responses between the baseline model (Llama-3.2-1B) and the teacher model used to generate labels for distillation (Mistral-7B). We provide  more details in Appendix~\ref{app:ragchecker}.

The model tuned with regular (short) labels generates shorter responses on average, hence it exhibits consistent natural drops in metrics such as \textit{recall} and \textit{context utilization}, which measure percentages of ground-truth claims or context claims entailed by the generated response. Retrieval metrics (claim recall, context precision) are the same across three models, since they all have the same retrieval setting. Self-knowledge measures a percentage of correct (i.e. entailed by ground truth) generated claims, not entailed by the retrieved context. We observe a zero value for self-knowledge for all models and datasets, showcasing that all three models generate responses by looking at the context rather than using internal LLM memory.

For the BioASQ dataset, which was evaluated with short labels, the model fine-tuned with short labels naturally achieves better RAGChecker results than the two other models.

\section{Conclusion}
This work addresses the problem of RAG adaptation under domain shift. In order to study this phenomenon we first build a large and diverse multi-domain benchmark for RAG. We then demonstrate that LLM finetuning for RAG with sequence-level distillation allows the model to better leverage the context, and transfers this capacity to other domains. 
\clearpage

\bibliography{latex/custom}
\bibliographystyle{ieeenat_fullname}
\clearpage
\appendix

\section{Datasets}

\begin{table}[ht]
    \centering
    \begin{tabular}{c|c}
        \textbf{Dataset} & \textbf{Number of queries} \\
        \hline
        HotpotQA \cite{yang2018hotpotqadatasetdiverseexplainable} & 88839\\
        NQ-open & 87925\\
        SQuAD \cite{rajpurkar2016squad100000questionsmachine} & 87596\\
        TriviaQA \cite{joshi2017triviaqalargescaledistantly} & 61797\\
        MSmarco & 59699\\
        AdversarialQA \cite{Bartolo_2020} & 29966\\
        FreebaseQA \cite{jiang-etal-2019-freebaseqa} & 20356\\
        SciQ \cite{welbl2017crowdsourcingmultiplechoicescience} & 11679\\
        ASQA \cite{stelmakh2023asqafactoidquestionsmeet} & 4353\\
        WikiQA \cite{yang-etal-2015-wikiqa} & 813\\
    \end{tabular}
    \caption{Content of MultiQA union of datasets}
    \label{tab:multiqa-content}
\end{table}

\section{Experimental details}
\label{appendix:ft_hyperparameters}

In Section \ref{sec:ragadapt}, we fine-tune models on general domain RAG data, to see if it improves performances on the out-of-domain RAG pipeline. Hyper-parameters are given in Table \ref{table:ft_hyperparameters}.

\begin{table}[h]
    \centering
    \begin{tabular}{|ll|}
    \hline
    \textbf{Hyperparameter} & \textbf{Value} \\ \hline
    Batch Size (Llama32-1B) & 512 \\
    Batch size (Llama3-8B) & 256 \\
    LR & $5 \times 10^{-4}$ \\
    LR scheduler & linear \\
    optimizer & AdamW \\
    Epochs & 1 \\
    LoRA Dropout & 0.1 \\ 
    LoRA Alpha & 32 \\ 
    Weight Decay & 0.1 \\ 
    Warmup Ratio & 0.05 \\ 
    Max Gradient Norm & 1.0 \\ 
    Documents max tokens & 128 \\
    \hline
    \end{tabular}
    \caption{Fine-tuning Hyperparameters.}
    \label{table:ft_hyperparameters}
\end{table}

\begin{table}[h]
    \centering
    \begin{tabular}{c|cc}
         Model & Rank &  Trainable parameters  \\
         \hline
         & \multicolumn{2}{c}{Llama3.2-1B}\\\hline
         FullLoRA &  16 & 11272192 \\
         LoRA-QKAtt & 128 & 13631488 \\
         LoRAAtt& 64 & 13631488 \\
         LoRAMLP & 24 & 11796480 \\\hline
          & \multicolumn{2}{c}{Llama3-8B}\\\hline
         FullLoRA &  16 & 41943040 \\
         LoRA-QKAtt & 128 &54525952 \\
         LoRAAtt& 64 & 54525952 \\
         LoRAMLP & 24 & 42467328 \\\hline

    \end{tabular}
    \caption{Rank and amount of trainable parameters for different LoRA variants. }
    \label{tab:lora_hyperparameters}
\end{table}

\section{Retrievers, Rerankers, and Top-k Retrieved Documents}

\begin{figure}[h!]
    \centering
    \includegraphics[width=\linewidth]{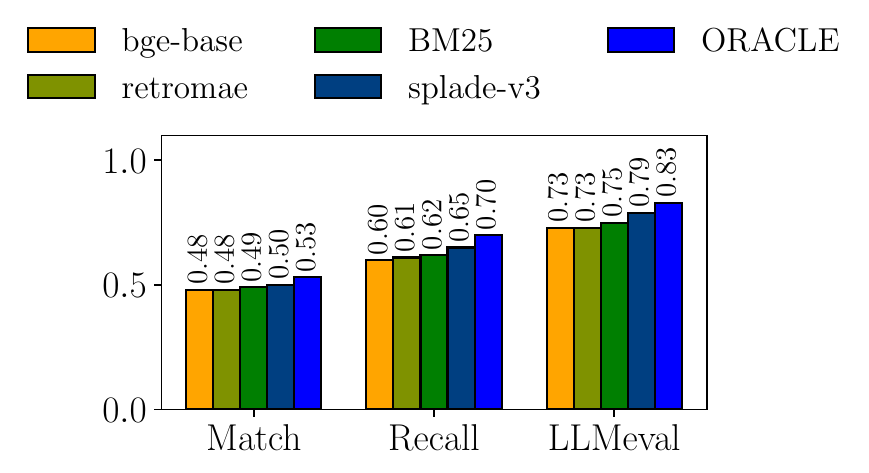}
    \caption{BioASQ-11b Performance Across Retrievers (SOLAR-10.7B generator)}
    \label{fig:retrievers-bioasq}
\end{figure}

\begin{figure}[h!]
    \centering
    \includegraphics[width=\linewidth]{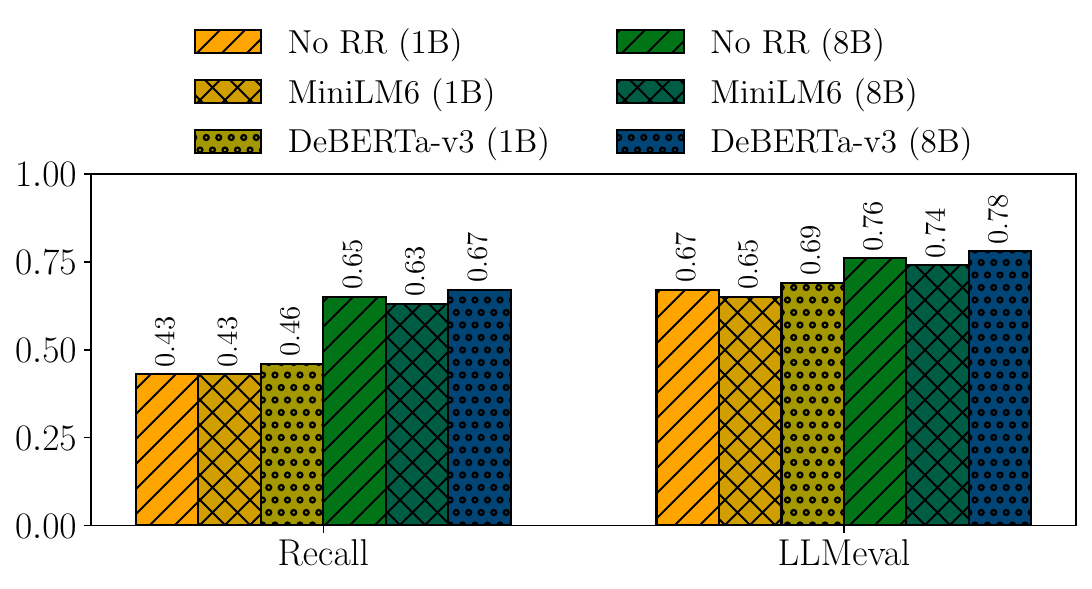}
    \caption{Reranking (RR) on BioASQ-12b dataset with splade-v3 retriever, with two generator of varying sizes (Llama3.2-1B-Instruct + Llama3-8B-Instruct) and comparing rerankers. Reranking with DeBERTa-v3 brings some improvement in RAG.}
    \label{fig:bioasq-reranker}
\end{figure}

In Figure \ref{fig:bioasq-top-k}, we highlight the importance of including not too many documents as this introduces noise for the generator, and not too few documents as this removes useful context for the generator. Using five documents seems like a good balance point. As shown in Figure \ref{fig:retrievers-bioasq}, we observe better performance when using oracle retrieval instead of top-5 retrieved documents. We can therefore say that suboptimal retrievers are unable to rank documents in the most optimal way to maximise generator performance, and/or generators are sensitive to surrounding noise around the useful context. We also notice the superior performance of the SPLADE-v3 retriever on the BioASQ task compared to other document retrievers. Finally, figure \ref{fig:bioasq-reranker} recalls the importance of rerankers for RAG performance. 

\section{Prompts}
\label{sec:appendix-reproducibility}

For each domain, we use simple domain-specific prompts to describe the task, which varies across datasets  especially across short-form and long-form QA. The generation temperature is set to $0$ for reproducibility. For gemma, qwen and llama models we use 4-bit quantization. For SOLAR-10.7B-Instruct we use vllm \cite{kwon2023efficient}).

\section{LLM evaluation}

\label{appendix:llm_eval}

\begin{table*}[h!]
\centering
\caption{Comparison between Match and LLMEval metrics across different examples.}
\renewcommand{\arraystretch}{1.5}
\begin{tabular}{|p{0.12\linewidth}|p{0.12\linewidth}|p{0.30\linewidth}|p{0.15\linewidth}|c|c|}
\hline
\textbf{Question} & \textbf{Gold Label} & \textbf{Model Output} & \textbf{Notes} & \textbf{Match} & \textbf{LLMEval} \\
\hline

Is there an association between bruxism and reflux? 
& yes 
& Based on the provided documents, it appears that there is some evidence suggesting a relationship between bruxism and gastroesophageal reflux disease (GERD). [...Analysis of provided RAG documents...] 
& LLMEval captures semantics better than Match (similar cases are frequent). 
& 0.0 & 1.0 \\
\hline

Which topoisomerase is essential in yeast? 
& topoisomerase II \textit{or} topo II 
& Based on the provided documents, there is a link between BCL11B haploinsufficiency and syndromic neurodevelopmental delay. [...explanation based on RAG documents...] In yeast, Topoisomerase I is essential, while Topoisomerase II is not essential. 
& LLMEval captures semantics better than Match metric. 
& 1.0 & 0.0 \\
\hline

Which is the database of somatic mutations in normal cells? 
& DSMNC \textit{or} Database of Somatic Mutations in Normal Cells 
& The correct answer is: Document 1: DSMNC: a database of somatic mutations in normal cells. This is the database of somatic mutations in normal cells, as mentioned in Document 2: SomaMutDB: a database of somatic mutations in normal human tissues. 
& LLMEval can be biased towards "final answer" (rare failure case). 
& 1.0 & 0.0 \\
\hline
\end{tabular}
\label{tab:llmeval_vs_match}

\end{table*}

To evaluate the quality of responses, we rely on an evaluation computed by a large language model with the prompt described in Figure \ref{fig:llm_eval_prompt}. Unless otherwise specified, we use the SOLAR-10.7B model\footnote{\href{https://huggingface.co/upstage/SOLAR-10.7B-Instruct-v1.0}{huggingface/upstage/SOLAR-10.7B-Instruct-v1.0}} as judge. \cite{rau2024bergenbenchmarkinglibraryretrievalaugmented} find that this metric has high correlation with GPT4. LLMEval better captures semantic meaning of text compared to static metrics like match (Figure \ref{tab:llmeval_vs_match}). 

\begin{figure}[h]
    \captionsetup{justification=centering}
    \caption{LLM Evaluation Prompt}
    \begin{mdframed}[backgroundcolor=gray!5, linewidth=0.8pt, linecolor=gray!75, innermargin=10pt, innerleftmargin=10pt, innerrightmargin=10pt]
        \textbf{system}: "You are an evaluation tool. Answer with one of 
            1: Correct,
            0.5: Partially correct,
            0: wrong.

        \textbf{user}: "Here is a question, a golden answer, and an AI-generated answer. Can you judge whether the AI-generated answer is correct according to the question and golden answer? Simply answer with one of 1: correct, 0.5: partially correct, 0: wrong. Question: \texttt{\{question\}}. Golden answer: \texttt{\{answer\}}. Generated answer: \texttt{\{prediction\}}."
    \end{mdframed}
    \label{fig:llm_eval_prompt}
\end{figure}

Table \ref{tab:llmeval_vs_match} provides several examples demonstrating that LLMeval is generally better suited to judge the correctness of the response compared to commonly used Match metric. 

\section{Other benchmark results}

Tables \ref{tab:benchmark-results-gemma-qwen},\ref{tab:benchmark-results-llama-solar},\ref{tab:benchmark-results-llama1B-llama8B} report additional benchmark results for small and medium-sized LLMs in terms of Recall and LLMEval.

\begin{table*}[hbt!]
    \begin{subtable}{\textwidth}
    \centering
    \captionsetup{justification=centering}
    \begin{tabular}{|l|cccc|cccc|}
        \hline
        \textbf{Dataset} & \multicolumn{4}{c|}{\textbf{gemma-2b-it}} & \multicolumn{4}{c|}{\textbf{qwen-2.5-3b-instruct}} \\
        \hline
         & \multicolumn{2}{c|}{\textbf{Without RAG}} & \multicolumn{2}{c|}{\textbf{With RAG}} & \multicolumn{2}{c|}{\textbf{Without RAG}} & \multicolumn{2}{c|}{\textbf{With RAG}} \\
        \hline
         & \textbf{Recall} & \textbf{LLMEval} & \textbf{Recall} & \textbf{LLMEval} & \textbf{Recall} & \textbf{LLMEval} & \textbf{Recall} & \textbf{LLMEval} \\
        \hline
        Bioasq11b & 0.341 & 0.357 & 0.470 & 0.589 & 0.382 & 0.491 & 0.649 & 0.769 \\
        Bioasq12b & 0.247 & 0.367 & 0.438 & 0.584 & 0.362 & 0.494 & 0.634 & 0.769 \\
        CovidQA
 & 0.229 & 0.213 & 0.357 & 0.453 & 0.305 & 0.336 & 0.480 & 0.559 \\
        FiQA
 & 0.170 & 0.308 & 0.169 & 0.317 & 0.200 & 0.406 & 0.201 & 0.391 \\
        ParaphraseRC
 & 0.104 & 0.104 & 0.211 & 0.322 & 0.190 & 0.141 & 0.553 & 0.572 \\
        Lifestyle
 & 0.234 & 0.241 & 0.201 & 0.320 & 0.304 & 0.390 & 0.337 & 0.565 \\
        Recreation & 0.192 & 0.140 & 0.185 & 0.261 & 0.299 & 0.238 & 0.381 & 0.498 \\
        Science & 0.280 & 0.288 & 0.242 & 0.376 & 0.325 & 0.454 & 0.339 & 0.457 \\
        Technology & 0.239 & 0.241 & 0.208 & 0.342 & 0.289 & 0.380 & 0.313 & 0.508 \\
        Writing & 0.231 & 0.257 & 0.193 & 0.313 & 0.335 & 0.470 & 0.353 & 0.549 \\
        SearchQA
 & 0.149 & 0.183 & 0.315 & 0.350 & 0.091 & 0.140 & 0.671 & 0.724 \\
        SyllabusQA
 & 0.256 & 0.300 & 0.195 & 0.231 & 0.339 & 0.312 & 0.391 & 0.278 \\
        TechQA & 0.237 & 0.184 & 0.268 & 0.253 & 0.274 & 0.256 & 0.412 & 0.528\\
        \hline
    \end{tabular}
    \caption{Benchmark Results for Small Models}
    \label{tab:benchmark-results-gemma-qwen}
    \end{subtable}

    \vspace*{0.5cm}

    \begin{subtable}{\textwidth}
    \centering
    \captionsetup{justification=centering}
    \begin{tabular}{|l|cccc|cccc|}
        \hline
        \textbf{Dataset} & \multicolumn{4}{c|}{\textbf{Llama3-8B-instruct}} & \multicolumn{4}{c|}{\textbf{SOLAR-10.7B-instruct}} \\
        \hline
         & \multicolumn{2}{c|}{\textbf{Without RAG}} & \multicolumn{2}{c|}{\textbf{With RAG}} & \multicolumn{2}{c|}{\textbf{Without RAG}} & \multicolumn{2}{c|}{\textbf{With RAG}} \\
        \hline
         & \textbf{Recall} & \textbf{LLMEval} & \textbf{Recall} & \textbf{LLMEval} & \textbf{Recall} & \textbf{LLMEval} & \textbf{Recall} & \textbf{LLMEval} \\
        \hline
        Bioasq11b & 0.446 & 0.595 & 0.615 & 0.762 & 0.445 & 0.622 & 0.668 & 0.791 \\
        Bioasq12b & 0.462 & 0.600 & 0.617 & 0.763 & 0.431 & 0.609 & 0.674 & 0.782 \\
        CovidQA
 & 0.321 & 0.328 & 0.501 & 0.551 & 0.311 & 0.405 & 0.503 & 0.605 \\
        FiQA
 & 0.198 & 0.432 & 0.230 & 0.438 & 0.196 & 0.492 & 0.218 & 0.499 \\
        ParaphraseRC
 & 0.274 & 0.279 & 0.607 & 0.624 & 0.377 & 0.356 & 0.635 & 0.648 \\
        Lifestyle
 & 0.299 & 0.442 & 0.358 & 0.592 & 0.301 & 0.549 & 0.360 & 0.688 \\
        Recreation & 0.282 & 0.283 & 0.367 & 0.532 & 0.308 & 0.361 & 0.412 & 0.603 \\
        Science & 0.320 & 0.464 & 0.364 & 0.568 & 0.325 & 0.426 & 0.377 & 0.634 \\
        Technology & 0.290 & 0.440 & 0.332 & 0.586 & 0.283 & 0.422 & 0.341 & 0.637 \\
        Writing & 0.287 & 0.521 & 0.363 & 0.629 & 0.318 & 0.567 & 0.386 & 0.740 \\
        SearchQA
 & 0.662 & 0.668 & 0.722 & 0.780 & 0.397 & 0.552 & 0.687 & 0.754 \\
        SyllabusQA
 & 0.382 & 0.304 & 0.311 & 0.313 & 0.309 & 0.295 & 0.305 & 0.284 \\
        TechQA & 0.288 & 0.266 & 0.469 & 0.576 & 0.273 & 0.300 & 0.461 & 0.597\\
        \hline
    \end{tabular}
    \caption{Benchmark Results for Medium-Sized Models}
    \label{tab:benchmark-results-llama-solar}
    \end{subtable}
    \caption{Benchmarks using models of different sizes (LLMEval is performed with Llama-3.1-70B\protect\footnotemark and a prompt specified in the Appendix \ref{sec:appendix-reproducibility}). The RAG pipeline uses \texttt{SPLADE-v3} retriever and \texttt{DeBERTa-v3} reranker.}
\end{table*}
\footnotetext{\url{https://huggingface.co/meta-llama/Llama-3.1-70B}}

\begin{table*}[hbt!]
    \begin{subtable}{\textwidth}
    \centering
    \captionsetup{justification=centering}
    \begin{tabular}{l|l|cccc|cccc}
       \multirow{3}{*}{\textbf{Domain }}& \multirow{3}{*}{\textbf{Dataset}} & \multicolumn{4}{c|}{\textbf{Llama-3.2-1B-Instruct}} & \multicolumn{4}{c}{\textbf{Llama-3-8B-Instruct}} \\
       
         & &  \multicolumn{2}{c}{\textbf{Without RAG}} & \multicolumn{2}{c|}{\textbf{With RAG}} & \multicolumn{2}{c}{\textbf{Without RAG}} & \multicolumn{2}{c}{\textbf{With RAG}} \\
         & &  \textbf{\small{Recall}} & \textbf{\small{LLMEval}} & \textbf{\small{Recall}} & \textbf{\small{LLMEval}} & \textbf{\small{Recall}} & \textbf{\small{LLMEval}} & \textbf{\small{Recall}} & \textbf{\small{LLMEval}} \\
        \hline
        \multirow{3}{*}{Biomedical}& Bioasq12b & 0.26 & 0.48 & 0.52 & 0.68 & 0.46 & 0.64 & 0.62 & 0.76 \\
        & CovidQA & 0.26 & 0.38 & 0.42 & 0.50 & 0.32 & 0.46 & 0.50 & 0.61 \\ 
        & \textbf{Mean}  & 0.26 & 0.43 & 0.47 & 0.59 & 0.39 & 0.55 & 0.56 &  0.69\\\hline
        \multirow{7}{*}{Long-form}& FiQA & 0.19 & 0.54 & 0.21 & 0.50 & 0.20 & 0.55 & 0.23 & 0.51 \\
        & Lifestyle & 0.27 & 0.50 & 0.32 & 0.56 & 0.30 & 0.51 & 0.36 & 0.62 \\
        & Recreation & 0.25 & 0.38 & 0.30 & 0.45 & 0.28 & 0.47 & 0.37 & 0.58 \\
        & Science & 0.31 & 0.49 & 0.34 & 0.49 & 0.32 & 0.33 & 0.36 & 0.64 \\
        & Technology & 0.27 & 0.47 & 0.30 & 0.53 & 0.29 & 0.43 & 0.33 & 0.64 \\
        & Writing & 0.25 & 0.44 & 0.32 & 0.49 & 0.29 & 0.53 & 0.36 & 0.67 \\
        & \textbf{Mean} & 0.26 & 0.47 & 0.30 & 0.50 & 0.28 & 0.47 & 0.34 & 0.61 \\ \hline
        Web-Search & SearchQA & 0.34 & 0.38 & 0.46 & 0.57 & 0.47 & 0.55 & 0.72 & 0.75 \\\hline
         \multirow{4}{*}{Context-critical}& ParaphraseRC & 0.16 & 0.18 & 0.44 & 0.47 & 0.27 & 0.32 & 0.61 & 0.63 \\
        & SyllabusQA & 0.27 & 0.37 & 0.28 & 0.30 & 0.38 & 0.40 & 0.31 & 0.26 \\
        & TechQA & 0.26 & 0.52 & 0.34 & 0.54 & 0.29 & 0.49 & 0.47 & 0.59 \\
        & \textbf{Mean} & 0.23 & 0.36 & 0.35 & 0.44 & 0.32 & 0.40 & 0.46 & 0.49 \\ 
        \hline
    \end{tabular}
    \end{subtable}
    \caption{Benchmarks using models of different sizes (LLMEval is performed with SOLAR-10.7B-Instruct\protect\footnotemark and a prompt specified in the Appendix \ref{sec:appendix-reproducibility}). The RAG pipeline uses \texttt{SPLADE-v3} retriever and \texttt{DeBERTa-v3} reranker.}
    \label{tab:benchmark-results-llama1B-llama8B}
\end{table*}
\footnotetext{\url{https://huggingface.co/upstage/SOLAR-10.7B-Instruct-v1.0}}

\begin{table*}[ht]
\centering
\resizebox{\textwidth}{!}{
\begin{tabular}{|c|c|c|c|c|c|c|c|c|c|c|}
\hline
 & \multirow{3}{*}{\textbf{Retriever}} & \multicolumn{3}{c|}{\textbf{Match}} & \multicolumn{3}{c|}{\textbf{Recall}} & \multicolumn{3}{c|}{\textbf{LLMeval}} \\ \cline{3-11}
 & & \textbf{Llm-1B} & \textbf{Llm-8B} & \textbf{Mstrl-7B} & \textbf{Llm-1B} & \textbf{Llm-8B} & \textbf{Mstrl-7B} & \textbf{Llm-1B} & \textbf{Llm-8B} & \textbf{Mstrl-7B} \\ \hline
\multirow{9}{*}{\rotatebox{90}{ KILT-NQ }} & bge & 0.536 & 0.623 & 0.654 & 0.624 & 0.702 & 0.736 & 0.567 & 0.682 & 0.712\\
& bge + RR & 0.562 & 0.654 & 0.68 & 0.655 & 0.733 & 0.766 & 0.593 & 0.705 & 0.731\\
& retromae & 0.52 & 0.597 & 0.62 & 0.604 & 0.677 & 0.702 & 0.536 & 0.652 & 0.68\\
& retromae + RR & 0.544 & 0.63 & 0.666 & 0.642 & 0.711 & 0.75 & 0.571 & 0.689 & 0.71\\
& bm25 & 0.392 & 0.451 & 0.479 & 0.483 & 0.542 & 0.574 & 0.439 & 0.531 & 0.572\\
& bm25 + RR & 0.459 & 0.556 & 0.579 & 0.562 & 0.64 & 0.673 & 0.506 & 0.619 & 0.649\\
& splade & 0.533 & 0.628 & 0.654 & 0.626 & 0.714 & 0.737 & 0.58 & 0.685 & 0.707\\
& splade + RR & 0.556 & 0.649 & 0.683 & 0.652 & 0.731 & 0.767 & 0.597 & 0.696 & 0.724\\
& oracle & 0.677 & 0.808 & 0.83 & 0.754 & 0.858 & 0.886 & 0.731 & 0.827 & 0.834\\
\hline
\multirow{9}{*}{\rotatebox{90}{ BioASQ-12b }} & bge & 0.302 & 0.416 & 0.443 & 0.428 & 0.545 & 0.566 & 0.621 & 0.705 & 0.718\\
& bge + RR & 0.33 & 0.462 & 0.502 & 0.471 & 0.608 & 0.636 & 0.675 & 0.742 & 0.757\\
& retromae & 0.288 & 0.42 & 0.468 & 0.421 & 0.551 & 0.612 & 0.615 & 0.698 & 0.724\\
& retromae + RR & 0.334 & 0.457 & 0.495 & 0.476 & 0.605 & 0.637 & 0.662 & 0.737 & 0.764\\
& bm25 & 0.287 & 0.39 & 0.437 & 0.393 & 0.503 & 0.56 & 0.584 & 0.675 & 0.704\\
& bm25 + RR & 0.318 & 0.418 & 0.473 & 0.446 & 0.553 & 0.61 & 0.62 & 0.701 & 0.723\\
& splade & 0.414 & 0.489 & 0.497 & 0.585 & 0.644 & 0.645 & 0.694 & 0.763 & 0.774\\
& splade + RR & 0.317 & 0.474 & 0.51 & 0.464 & 0.634 & 0.654 & 0.682 & 0.775 & 0.783\\
& oracle & 0.351 & 0.468 & 0.517 & 0.524 & 0.633 & 0.672 & 0.69 & 0.78 & 0.794\\
\hline
\multirow{9}{*}{\rotatebox{90}{ TechQA }} & bge & 0.014 & 0.066 & 0.024 & 0.35 & 0.49 & 0.44 & 0.554 & 0.605 & 0.632\\
& bge + RR & 0.014 & 0.066 & 0.024 & 0.35 & 0.47 & 0.432 & 0.547 & 0.596 & 0.623\\
& retromae & 0.011 & 0.055 & 0.018 & 0.339 & 0.468 & 0.421 & 0.529 & 0.617 & 0.61\\
& retromae + RR & 0.011 & 0.06 & 0.019 & 0.34 & 0.464 & 0.424 & 0.52 & 0.607 & 0.605\\
& bm25 & 0.018 & 0.056 & 0.016 & 0.353 & 0.468 & 0.418 & 0.514 & 0.598 & 0.608\\
& bm25 + RR & 0.016 & 0.052 & 0.019 & 0.345 & 0.469 & 0.424 & 0.527 & 0.598 & 0.596\\
& splade & 0.01 & 0.042 & 0.021 & 0.346 & 0.443 & 0.442 & 0.528 & 0.635 & 0.621\\
& splade + RR & 0.014 & 0.06 & 0.023 & 0.356 & 0.475 & 0.423 & 0.539 & 0.601 & 0.609\\
& oracle & - & - & - & - & - & - & - & - & \\
\hline
\multirow{9}{*}{\rotatebox{90}{ FiQA }} & bge & 0.0 & 0.001 & 0.0 & 0.208 & 0.233 & 0.226 & 0.506 & 0.539 & 0.582\\
& bge + RR & 0.0 & 0.0 & 0 & 0.208 & 0.232 & 0.227 & 0.52 & 0.544 & 0.591\\
& retromae & 0.0 & 0.001 & 0 & 0.202 & 0.226 & 0.22 & 0.479 & 0.521 & 0.575\\
& retromae + RR & 0.001 & 0.0 & 0 & 0.207 & 0.231 & 0.224 & 0.5 & 0.544 & 0.594\\
& bm25 & 0.0 & 0 & 0 & 0.188 & 0.216 & 0.212 & 0.399 & 0.485 & 0.562\\
& bm25 + RR & 0.0 & 0.001 & 0 & 0.2 & 0.226 & 0.219 & 0.456 & 0.52 & 0.575\\
& splade & 0 & 0 & 0 & 0.189 & 0.182 & 0.22 & 0.485 & 0.563 & 0.586\\
& splade + RR & 0.001 & 0.0 & 0 & 0.208 & 0.231 & 0.225 & 0.523 & 0.545 & 0.59\\
& oracle & - & - & - & - & - & - & - & - & \\
\hline
\multirow{9}{*}{\rotatebox{90}{ SyllabusQA }} & bge & 0.056 & 0.057 & 0.071 & 0.271 & 0.305 & 0.347 & 0.294 & 0.272 & 0.349\\
& bge + RR & 0.057 & 0.057 & 0.076 & 0.285 & 0.314 & 0.366 & 0.301 & 0.289 & 0.357\\
& retromae & 0.061 & 0.05 & 0.072 & 0.276 & 0.296 & 0.35 & 0.293 & 0.255 & 0.372\\
& retromae + RR & 0.057 & 0.056 & 0.068 & 0.277 & 0.312 & 0.354 & 0.298 & 0.284 & 0.374\\
& bm25 & 0.087 & 0.106 & 0.147 & 0.361 & 0.425 & 0.502 & 0.443 & 0.555 & 0.565\\
& bm25 + RR & 0.082 & 0.114 & 0.153 & 0.37 & 0.442 & 0.522 & 0.478 & 0.608 & 0.604\\
& splade & 0.057 & 0.061 & 0.066 & 0.266 & 0.272 & 0.351 & 0.237 & 0.28 & 0.33\\
& splade + RR & 0.057 & 0.06 & 0.075 & 0.278 & 0.305 & 0.363 & 0.322 & 0.289 & 0.342\\
& oracle & - & - & - & - & - & - & - & - & \\
\hline

\end{tabular}}
\caption{Three metrics across 5 datasets, 3 generators, 5 retrievers and reranking (RR).}
\label{tab:fig1_full_table}
\end{table*}

\begin{table*}
\centering

\begin{tabular}{l|l l l l}
 & \multicolumn{2}{c}{Llama-3.2-1B}& \multicolumn{2}{c}{Llama-3-8B}\\
 & KILT-NQ & BioASQ12b & KILT-NQ & BioASQ12b \\\hline
Vanilla RAG & 0.75 & 0.52 & 0.86 & 0.63 \\\hline
 & \multicolumn{4}{c}{FT with LoRA }\\\hline
Full-LoRA & 0.71 & 0.41 & 0.76 & 0.36 \\\hline
LoRA-QKAtt & 0.68 & 0.58 & 0.78 & 0.53 \\\hline
LoRA-Att & 0.7 & 0.43 & 0.78 & 0.41 \\\hline
LoRA-MLP & 0.71 & 0.43 & 0.75 & 0.37 \\\hline
 & \multicolumn{4}{c}{FT with LoRA + distilled labels}\\\hline
Full-LoRA & 0.87 & 0.67 & 0.89 & 0.69 \\\hline

\end{tabular}
\caption{Recall scores for RAG with oracle documents in the context,  with and without FineTuning for RAG. }
\label{tab:recall_oracle}
\end{table*}

\section{Training with distractors.}
\label{app:distractors}
In addition to the main methods,  we experiment with previously proposed strategies \cite{zhang2024raftadaptinglanguagemodel}, that add noise in the retrieved documents during training, in order to make models more resilient to bad context. To do so, after retrieval and reranking is performed, we replace the least $D$ relevant documents with documents sampled at random from the documents repository.

Table \ref{tab:distractors} reports the results of training with distractors. We note that, as expected, \textbf{training with distractors does not necessarily make the model more robust for new domains.} We explore robustness to the context more in depth in the Appendix \ref{app:robustness}. 

\begin{table*}[!htbp]
    \centering
    \begin{tabular}{c|cccccc}
         Setup & General & Biomed. & WebSearch & CtxtCritical & LongForm & Avg \\
         \hline
         Full-LoRA & 0.65 & 0.63 & 0.68 & 0.41 & 0.42 & 0.50\\
         Full-LoRA + Distractors & 0.61 & 0.62 & 0.67 & 0.40 & 0.41 & 0.49 \\
         LoRA-QKAtt + Distractors & 0.60 & 0.62 & 0.59 & 0.43 & 0.45 & 0.50 \\\hline
         Full-LoRA distill & 0.64 & 0.67 & 0.70 & 0.49 & 0.61 & 0.58\\
         Full-LoRA distill + Distractors & 0.60 & 0.66 & 0.69 & 0.46 & 0.59 & 0.59\\\hline
    \end{tabular}
    \caption{Training with distractors, Llama-3.2-1B}
    \label{tab:distractors}
\end{table*}

\section{Context robustness analysis.}
\label{app:robustness}
We investigate how do RAG adaptation techniques impact model robustness to noisy context. 
 Similar to \cite{Niu2020} we introduce robustness metric $R(y, y')$ measuring the ratio of correct answers produced by RAG model dealing with noisy context in comparison to the same model dealing with clean context. Clean context refers to our default RAG settings (splade-v3 retriever, DeBERTa-v3 reranker, top5 documents). We compare 3 types of  noisy context: (1) randomly replace 4 out of 5 relevant documents by distractors (2) remove reranking from the pipeline (3) extend top-k documents (from 5 to 20).

 Table \ref{tab:robustness} reports the results of robustness evaluation across 4 datasets belonging to different domains (NQ, Bioasq12b, TechQA and FiQA). In addition to the robustness metric we report LLMEval performance obtained in clean settings. The best model should have highest LLMEval in clean settings, and highest robustness scores  (e.g. being resilient to the distribution shift in the way the relevant information is presented). 

First, we note that, similar to the observation made in section \ref{sec:benchmark} (Figure \ref{fig:bioasq-top-k}), the smaller model (\textbf{Llama-3.2-1B}) 
is more sensitive to noisy context compared to \textbf{Llama-3-8B}.

 We also note that models trained with distractors are indeed more robust when the same type of noise is present at test time, but not necessarily to other  types of distribution shift (top-20, no reranking) when trained with regular labels.  Overall, original model (Lama-3.2-1B) seems to be more robust to the distribution shift in the relevant information, compared to all finetuned variants (especially on general domain). We attribute it to strong capabilities of instruction-tuned LLMs to exploit their context. However, highest robustness does not necessarily correlate with the best  performance in clean settings.  
 Models trained with distilled labels and distractors seems to represent the best trade-off between robustness and overall performance. This aligns with our hypothesis, that such a model diverges less from the original distribution and focuses more on better exploiting relevant information.

\begin{table*}
\centering

\begin{tabular}{ l | l  l  l | c | l  l  l | c }
\toprule
 & \multicolumn{4}{c}{General domain} & \multicolumn{4}{c}{Other domains}\\
 & Distr. & NoRR & Top20 & LLMEval & Distr. & No RR & Top20 & LLMEval \\

\midrule
Llama3-8B & 0.89 & 0.97 & 1.02 & 0.65 & 0.93 & 1.03 & 1.05 & 0.64 \\
\hline
Llama32-1B &\textbf{0.96} & \textbf{0.96} & \textbf{0.99} & 0.53 & 0.88 & 0.96 & \textbf{1.00} & 0.58 \\
\hline
FullLoRA & 0.83 & 0.93 & 0.95 & 0.65 & 0.91 & 0.96 & 0.96 & 0.49 \\
LoRAQKAtt & 0.88 & 0.92 & 0.91 & 0.62 & 0.88 & 0.90 & 0.92 & 0.53  \\

LoRAAtt & 0.84 & 0.92 & 0.93 & 0.65 & 0.97 & \textbf{1.00} & \textbf{1.00} & 0.48  \\
LoRAMLP& 0.84 & 0.92 & 0.94 & \textbf{0.66} & 0.95 & 0.98 & 0.98 & 0.49 \\
FullLoRA + distilled labels & 0.87 & 0.93 & 0.95 & 0.64 & 0.83 & 0.93 & 0.96 & \textbf{0.63} \\
\hline
&\multicolumn{8}{|c}{ + distractors} \\\hline
LoRAQKAtt & 0.93 & 0.91 & 0.92 & 0.60 & 0.92 & 0.92 & 0.94 & 0.53\\
FullLoRA & 0.94 & 0.95 & 0.98 & 0.61 & \textbf{0.98} & 0.99 & 0.97 & 0.48  \\
FullLoRA + distilled labels & \textbf{0.98} & 0.94 & 0.91 & 0.60 & 0.94 & 0.94 & 0.91 & 0.61 \\
\hline

\end{tabular}
\caption{Robustness metric for different types of noise (1) distractors (2) No reranking (3) top 20; for different domains (1) General (NQ and PopQA) (2) Other domains (Bioasq12b, TechQA, FiQA).}
\label{tab:robustness}
\end{table*}

\section{When does RAG fail?}

A well known struggle in RAG is the ability of models to adapt to contextual knowledge. Context can either clash with the model's internal knowledge \cite{monea2024glitchmatrixlocatingdetecting}, or, in the case of multidomain RAG, the model may have no internal knowledge at all. Figure \ref{fig:rag_fail}  gives a formal example in \ref{fig:rag_fail}, where the model struggles to adapt to out-of-distribution text. 

\begin{figure}[H]
    \captionsetup{justification=centering}
    \caption{Generator fails to exploit context in the biomedical domain.}
    \begin{mdframed}[backgroundcolor=gray!5, linewidth=0.8pt, linecolor=gray!75, innermargin=10pt, innerleftmargin=10pt, innerrightmargin=10pt]
        \textbf{Background:}\newline Document 1: [...] we found that the predominantly \textcolor{cyan}{nuclear} A3\textcolor{cyan}{B} [...] In contrast to the exclusively \textcolor{red}{cytoplasmic} A3\textcolor{red}{G} \newline
        Document 2: [...] APOBEC3\textcolor{red}{G} limits the replication of [...]. It localizes predominantly to the \textcolor{red}{cytoplasm} of cells [...]\\
        Document 3: [...]\\
        \textbf{Question:} Is APOBEC3\textcolor{cyan}{B} protein predominantly cytoplasmic or nuclear?\\
        \textbf{Response:} Based on the provided documents, it appears that the APOBEC3\textcolor{cyan}{B} protein is predominantly \textcolor{red}{cytoplasmic}. The documents mention that:\newline
        - APOBEC3\textcolor{cyan}{B} is predominantly nuclear in the cytoplasm of cells.\newline
        - APOBEC3\textcolor{cyan}{B} localizes to the cytoplasm of cells, which is consistent with a model where it packages into assembling virions.\\
        \textbf{Label: \textcolor{cyan}{nuclear}}
    \end{mdframed}
    \label{fig:rag_fail}
\end{figure}

\section{RAGChecker's sensitivity to style of the LLM response}
\label{app:ragchecker}
In manual inspection of RAGChecker results in Section~\ref{sec:ragchecker}, we found that \textit{style difference between two models may cause difference in RAGChecker metrics}. 
Particularly, RAGChecker extracts atomic claims from a model-generated response and makes judgment about them depending on whether a claim is entailed by the ground truth answer or by the retrieved context. 

If two models differ in style, e.g. one model inserts generic comments into its reply more often than another, then these generic comments will not be entailed by the ground truth answer or by the retrieved context and will be treated \textit{the same way as wrong claims}. This may reduce the values of some RAGChecker metrics for one of the models.

In our case, \verb|Llama-3.2-1B| model trained on labels distilled from \verb|Mistral-7B| exhibits higher \textit{Faithfulness} and lower \textit{Hallucination} than a baseline  \verb|Llama-3.2-1B| model applied zero-shot. We found that the latter model tends to repeat the user's question more frequently than the former model, e.g. ``The question is asking what prevents interest rates from rising.''. We confirmed this quantitatively: the number of claims extracted from model responses, containing keywords ``question'' and ``ask'', is 96 for the baseline model and only 15 for the model trained with distilled labels. 
All these question repetitions will be counted as claims not entailed by the ground truth response or by the retrieved context, hence reducing \textit{Faithfulness}\footnote{Faithfulness is defined as a portion of response claims entailed by the retrieved context.} and increasing \textit{Hallucination}\footnote{Hallucination is defined as a portion of response claims not entailed by the ground truth answer and not entailed by the context}.

\section{Full tables for RAG adaptation}

In this section, we display all RAG adaptation results by individual datasets in Tables \ref{tab:rag_adaptation_full_general} - \ref{tab:rag_adaptation_full_avg}. 

\begin{table*}[!ht]
\centering
\begin{tabular}{|c|c|cccccc|}
\hline
 & \multirow{2}{*}{\textbf{Adaptation Type}} & \multicolumn{2}{c|}{\textbf{Match}} & \multicolumn{2}{c|}{\textbf{Recall}} & \multicolumn{2}{c|}{\textbf{LLMeval}} \\ \cline{3-8}
 & & \textbf{Llm-1B} & \textbf{Llm-8B} & \textbf{Llm-1B} & \textbf{Llm-8B} & \textbf{Llm-1B} & \textbf{Llm-8B} \\ \hline
\multirow{11}{*}{\rotatebox{90}{ KILT-NQ }} & Vanilla RAG & 0.557 & 0.698 & 0.651 & 0.729 & 0.6 & 0.698\\
& FullLora & 0.506 & 0.779 & 0.614 & 0.69 & 0.7 & 0.779\\
& LoraQKAtt & 0.506 & 0.756 & 0.57 & 0.654 & 0.67 & 0.756\\
& LoraAtt & 0.536 & 0.765 & 0.601 & 0.675 & 0.69 & 0.765\\
& LoraMLP & 0.558 & 0.776 & 0.618 & 0.684 & 0.71 & 0.776\\
\cline{2-8}
& FullLora-distill & 0.644 & 0.729 & 0.731 & 0.759 & 0.7 & 0.729\\
& LoraQKAtt-distill & 0.63 & 0.726 & 0.719 & 0.763 & 0.69 & 0.726\\
& LoraAtt-distill & 0.645 & 0.732 & 0.729 & 0.764 & 0.7 & 0.732\\
& LoraMLP-distill & 0.655 & 0.728 & 0.741 & 0.764 & 0.7 & 0.728\\
\cline{2-8}
& FullLora-distractors & 0.527 &  & 0.588 &  & 0.68 & \\
& LoraQKAtt-distractors & 0.49 &  & 0.555 &  & 0.66 & \\
& FullLora-distill-distractors & 0.617 &  & 0.703 &  & 0.68 & \\
\hline
\multirow{11}{*}{\rotatebox{90}{ PopQA }} & Vanilla RAG & 0.595 & 0.6 & 0.627 & 0.687 & 0.47 & 0.6\\
& FullLora & 0.516 & 0.642 & 0.578 & 0.609 & 0.61 & 0.642\\
& LoraQKAtt & 0.516 & 0.639 & 0.542 & 0.608 & 0.57 & 0.639\\
& LoraAtt & 0.545 & 0.641 & 0.574 & 0.611 & 0.6 & 0.641\\
& LoraMLP & 0.546 & 0.645 & 0.577 & 0.613 & 0.6 & 0.645\\
\cline{2-8}
& FullLora-distill & 0.661 & 0.586 & 0.679 & 0.708 & 0.57 & 0.586\\
& LoraQKAtt-distill & 0.629 & 0.601 & 0.646 & 0.711 & 0.58 & 0.601\\
& LoraAtt-distill & 0.652 & 0.59 & 0.67 & 0.712 & 0.58 & 0.59\\
& LoraMLP-distill & 0.66 & 0.589 & 0.678 & 0.709 & 0.57 & 0.589\\
\cline{2-8}
& FullLora-distractors & 0.478 &  & 0.509 &  & 0.53 & \\
& LoraQKAtt-distractors & 0.484 &  & 0.511 &  & 0.54 & \\
& FullLora-distill-distractors & 0.597 &  & 0.619 &  & 0.53 & \\
\hline
\multirow{11}{*}{\rotatebox{90}{ General }} & Vanilla RAG & 0.58 & 0.65 & 0.64 & 0.71 & 0.53 & 0.65\\
& FullLora & 0.55 & 0.71 & 0.6 & 0.65 & 0.65 & 0.71\\
& LoraQKAtt & 0.51 & 0.7 & 0.56 & 0.63 & 0.62 & 0.7\\
& LoraAtt & 0.54 & 0.7 & 0.59 & 0.64 & 0.65 & 0.7\\
& LoraMLP & 0.55 & 0.71 & 0.6 & 0.65 & 0.66 & 0.71\\
\cline{2-8}
& FullLora-distill & 0.65 & 0.66 & 0.71 & 0.73 & 0.64 & 0.66\\
& LoraQKAtt-distill & 0.63 & 0.66 & 0.68 & 0.74 & 0.63 & 0.66\\
& LoraAtt-distill & 0.65 & 0.66 & 0.7 & 0.74 & 0.64 & 0.66\\
& LoraMLP-distill & 0.66 & 0.66 & 0.71 & 0.74 & 0.64 & 0.66\\
\cline{2-8}
& FullLora-distractors & 0.5 &  & 0.55 &  & 0.61 & \\
& LoraQKAtt-distractors & 0.49 &  & 0.53 &  & 0.6 & \\
& FullLora-distill-distractors & 0.61 &  & 0.66 &  & 0.6 & \\
\hline
\end{tabular}
\caption{RAG Adaptation, General Domain}
\label{tab:rag_adaptation_full_general}
\end{table*}

\begin{table*}
\centering
\begin{tabular}{|c|c|cccccc|}
\hline
 & \multirow{2}{*}{\textbf{Adaptation Type}} & \multicolumn{2}{c|}{\textbf{Match}} & \multicolumn{2}{c|}{\textbf{Recall}} & \multicolumn{2}{c|}{\textbf{LLMeval}} \\ \cline{3-8}
 & & \textbf{Llm-1B} & \textbf{Llm-8B} & \textbf{Llm-1B} & \textbf{Llm-8B} & \textbf{Llm-1B} & \textbf{Llm-8B} \\ \hline
\multirow{11}{*}{\rotatebox{90}{ BioASQ-12b }} & Vanilla RAG & 0.295 & 0.762 & 0.432 & 0.614 & 0.67 & 0.762\\
& FullLora & 0.448 & 0.8 & 0.511 & 0.599 & 0.73 & 0.8\\
& LoraQKAtt & 0.448 & 0.764 & 0.566 & 0.499 & 0.71 & 0.764\\
& LoraAtt & 0.418 & 0.801 & 0.508 & 0.6 & 0.72 & 0.801\\
& LoraMLP & 0.423 & 0.793 & 0.523 & 0.589 & 0.73 & 0.793\\
\cline{2-8}
& FullLora-distill & 0.499 & 0.795 & 0.646 & 0.673 & 0.75 & 0.795\\
& LoraQKAtt-distill & 0.463 & 0.778 & 0.605 & 0.681 & 0.74 & 0.778\\
& LoraAtt-distill & 0.491 & 0.786 & 0.635 & 0.664 & 0.76 & 0.786\\
& LoraMLP-distill & 0.479 & 0.795 & 0.627 & 0.665 & 0.75 & 0.795\\
\cline{2-8}
& FullLora-distractors & 0.426 &  & 0.527 &  & 0.73 & \\
& LoraQKAtt-distractors & 0.429 &  & 0.557 &  & 0.72 & \\
& FullLora-distill-distractors & 0.509 &  & 0.644 &  & 0.75 & \\
\hline
\multirow{11}{*}{\rotatebox{90}{ CovidQA }} & Vanilla RAG & 0.131 & 0.614 & 0.429 & 0.487 & 0.51 & 0.614\\
& FullLora & 0.17 & 0.595 & 0.244 & 0.307 & 0.53 & 0.595\\
& LoraQKAtt & 0.17 & 0.595 & 0.325 & 0.371 & 0.55 & 0.595\\
& LoraAtt & 0.12 & 0.591 & 0.234 & 0.296 & 0.53 & 0.591\\
& LoraMLP & 0.117 & 0.591 & 0.232 & 0.295 & 0.53 & 0.591\\
\cline{2-8}
& FullLora-distill & 0.156 & 0.637 & 0.457 & 0.478 & 0.59 & 0.637\\
& LoraQKAtt-distill & 0.155 & 0.633 & 0.451 & 0.478 & 0.57 & 0.633\\
& LoraAtt-distill & 0.158 & 0.625 & 0.461 & 0.48 & 0.59 & 0.625\\
& LoraMLP-distill & 0.159 & 0.633 & 0.467 & 0.477 & 0.59 & 0.633\\
\cline{2-8}
& FullLora-distractors & 0.107 &  & 0.224 &  & 0.51 & \\
& LoraQKAtt-distractors & 0.138 &  & 0.281 &  & 0.52 & \\
& FullLora-distill-distractors & 0.122 &  & 0.418 &  & 0.57 & \\
\hline
\multirow{11}{*}{\rotatebox{90}{ Biomedical Avg }} & Vanilla RAG & 0.21 & 0.69 & 0.43 & 0.55 & 0.59 & 0.69\\
& FullLora & 0.27 & 0.7 & 0.38 & 0.45 & 0.63 & 0.7\\
& LoraQKAtt & 0.31 & 0.68 & 0.45 & 0.44 & 0.63 & 0.68\\
& LoraAtt & 0.27 & 0.7 & 0.37 & 0.45 & 0.62 & 0.7\\
& LoraMLP & 0.27 & 0.69 & 0.38 & 0.44 & 0.63 & 0.69\\
\cline{2-8}
& FullLora-distill & 0.33 & 0.72 & 0.55 & 0.58 & 0.67 & 0.72\\
& LoraQKAtt-distill & 0.31 & 0.71 & 0.53 & 0.58 & 0.66 & 0.71\\
& LoraAtt-distill & 0.32 & 0.71 & 0.55 & 0.57 & 0.67 & 0.71\\
& LoraMLP-distill & 0.32 & 0.71 & 0.55 & 0.57 & 0.67 & 0.71\\
\cline{2-8}
& FullLora-distractors & 0.27 &  & 0.38 &  & 0.62 & \\
& LoraQKAtt-distractors & 0.28 &  & 0.42 &  & 0.62 & \\
& FullLora-distill-distractors & 0.32 &  & 0.53 &  & 0.66 & \\
\hline
\end{tabular}
\caption{RAG Adaptation, Biomedical Domain. }
\label{tab:rag_adaptation_full_biomedical}
\end{table*}

\begin{table*}
\centering
\begin{tabular}{|c|c|cccccc|}
\hline
 & \multirow{2}{*}{\textbf{Adaptation Type}} & \multicolumn{2}{c|}{\textbf{Match}} & \multicolumn{2}{c|}{\textbf{Recall}} & \multicolumn{2}{c|}{\textbf{LLMeval}} \\ \cline{3-8}
 & & \textbf{Llm-1B} & \textbf{Llm-8B} & \textbf{Llm-1B} & \textbf{Llm-8B} & \textbf{Llm-1B} & \textbf{Llm-8B} \\ \hline
\multirow{11}{*}{\rotatebox{90}{ FiQA }} & Vanilla RAG & 0.0 & 0.51 & 0.201 & 0.23 & 0.49 & 0.51\\
& FullLora & 0.004 & 0.434 & 0.034 & 0.06 & 0.39 & 0.434\\
& LoraQKAtt & 0.004 & 0.522 & 0.105 & 0.102 & 0.44 & 0.522\\
& LoraAtt & 0 & 0.423 & 0.035 & 0.044 & 0.39 & 0.423\\
& LoraMLP & 0 & 0.442 & 0.031 & 0.056 & 0.37 & 0.442\\
\cline{2-8}
& FullLora-distill & 0 & 0.606 & 0.228 & 0.233 & 0.55 & 0.606\\
& LoraQKAtt-distill & 0 & 0.59 & 0.235 & 0.241 & 0.54 & 0.59\\
& LoraAtt-distill & 0 & 0.594 & 0.233 & 0.233 & 0.54 & 0.594\\
& LoraMLP-distill & 0 & 0.606 & 0.231 & 0.236 & 0.55 & 0.606\\
\cline{2-8}
& FullLora-distractors & 0 &  & 0.027 &  & 0.35 & \\
& LoraQKAtt-distractors & 0.002 &  & 0.077 &  & 0.42 & \\
& FullLora-distill-distractors & 0 &  & 0.212 &  & 0.53 & \\
\hline
\multirow{11}{*}{\rotatebox{90}{ RobustQA-Lifestyle }} & Vanilla RAG & 0.003 & 0.616 & 0.492 & 0.506 & 0.57 & 0.616\\
& FullLora & 0.004 & 0.496 & 0.041 & 0.063 & 0.43 & 0.496\\
& LoraQKAtt & 0.004 & 0.603 & 0.123 & 0.14 & 0.54 & 0.603\\
& LoraAtt & 0.002 & 0.476 & 0.041 & 0.047 & 0.42 & 0.476\\
& LoraMLP & 0.002 & 0.495 & 0.037 & 0.065 & 0.41 & 0.495\\
\cline{2-8}
& FullLora-distill & 0.002 & 0.661 & 0.331 & 0.359 & 0.62 & 0.661\\
& LoraQKAtt-distill & 0.002 & 0.659 & 0.389 & 0.397 & 0.6 & 0.659\\
& LoraAtt-distill & 0.002 & 0.653 & 0.35 & 0.365 & 0.6 & 0.653\\
& LoraMLP-distill & 0.002 & 0.662 & 0.336 & 0.362 & 0.63 & 0.662\\
\cline{2-8}
& FullLora-distractors & 0.002 &  & 0.042 &  & 0.43 & \\
& LoraQKAtt-distractors & 0.002 &  & 0.098 &  & 0.49 & \\
& FullLora-distill-distractors & 0.002 &  & 0.298 &  & 0.59 & \\
\hline
\multirow{11}{*}{\rotatebox{90}{ RobustQA-Recreation }} & Vanilla RAG & 0.0 & 0.584 & 0.43 & 0.404 & 0.47 & 0.584\\
& FullLora & 0.002 & 0.49 & 0.07 & 0.093 & 0.42 & 0.49\\
& LoraQKAtt & 0.002 & 0.55 & 0.126 & 0.142 & 0.48 & 0.55\\
& LoraAtt & 0.001 & 0.477 & 0.063 & 0.076 & 0.4 & 0.477\\
& LoraMLP & 0.001 & 0.492 & 0.065 & 0.096 & 0.41 & 0.492\\
\cline{2-8}
& FullLora-distill & 0.001 & 0.618 & 0.355 & 0.376 & 0.56 & 0.618\\
& LoraQKAtt-distill & 0.001 & 0.628 & 0.347 & 0.372 & 0.55 & 0.628\\
& LoraAtt-distill & 0.001 & 0.607 & 0.366 & 0.38 & 0.56 & 0.607\\
& LoraMLP-distill & 0.001 & 0.614 & 0.355 & 0.376 & 0.56 & 0.614\\
\cline{2-8}
& FullLora-distractors & 0.0 &  & 0.066 &  & 0.4 & \\
& LoraQKAtt-distractors & 0.0 &  & 0.1 &  & 0.44 & \\
& FullLora-distill-distractors & 0.001 &  & 0.316 &  & 0.52 & \\
\hline
\multirow{11}{*}{\rotatebox{90}{ RobustQA-Science }} & Vanilla RAG & 0.001 & 0.636 & 0.481 & 0.5 & 0.5 & 0.636\\
& FullLora & 0.004 & 0.484 & 0.065 & 0.074 & 0.42 & 0.484\\
& LoraQKAtt & 0.004 & 0.591 & 0.134 & 0.16 & 0.48 & 0.591\\
& LoraAtt & 0.001 & 0.47 & 0.058 & 0.061 & 0.4 & 0.47\\
& LoraMLP & 0.001 & 0.494 & 0.064 & 0.079 & 0.43 & 0.494\\
\cline{2-8}
& FullLora-distill & 0.003 & 0.693 & 0.361 & 0.376 & 0.64 & 0.693\\
& LoraQKAtt-distill & 0.003 & 0.669 & 0.377 & 0.407 & 0.62 & 0.669\\
& LoraAtt-distill & 0.001 & 0.681 & 0.37 & 0.382 & 0.63 & 0.681\\
& LoraMLP-distill & 0.003 & 0.696 & 0.362 & 0.382 & 0.64 & 0.696\\
\cline{2-8}
& FullLora-distractors & 0.001 &  & 0.074 &  & 0.43 & \\
& LoraQKAtt-distractors & 0.001 &  & 0.097 &  & 0.45 & \\
& FullLora-distill-distractors & 0.004 &  & 0.329 &  & 0.63 & \\
\hline
\end{tabular}
\caption{RAG Adaptation, Long-Form domain (1). }
\label{tab:rag_adaptation_full_longform1}
\end{table*}

\begin{table*}
\centering
\begin{tabular}{|c|c|cccccc|}
\hline
 & \multirow{2}{*}{\textbf{Adaptation Type}} & \multicolumn{2}{c|}{\textbf{Match}} & \multicolumn{2}{c|}{\textbf{Recall}} & \multicolumn{2}{c|}{\textbf{LLMeval}} \\ \cline{3-8}
 & & \textbf{Llm-1B} & \textbf{Llm-8B} & \textbf{Llm-1B} & \textbf{Llm-8B} & \textbf{Llm-1B} & \textbf{Llm-8B} \\ \hline
\multirow{11}{*}{\rotatebox{90}{ RobustQA-Technology }} & Vanilla RAG & 0.0 & 0.635 & 0.442 & 0.433 & 0.53 & 0.635\\
& FullLora & 0.003 & 0.49 & 0.068 & 0.083 & 0.43 & 0.49\\
& LoraQKAtt & 0.003 & 0.572 & 0.118 & 0.139 & 0.49 & 0.572\\
& LoraAtt & 0.002 & 0.472 & 0.063 & 0.067 & 0.41 & 0.472\\
& LoraMLP & 0.001 & 0.487 & 0.061 & 0.083 & 0.41 & 0.487\\
\cline{2-8}
& FullLora-distill & 0.0 & 0.707 & 0.319 & 0.336 & 0.63 & 0.707\\
& LoraQKAtt-distill & 0.001 & 0.693 & 0.347 & 0.353 & 0.61 & 0.693\\
& LoraAtt-distill & 0.0 & 0.697 & 0.333 & 0.339 & 0.62 & 0.697\\
& LoraMLP-distill & 0.001 & 0.705 & 0.323 & 0.34 & 0.63 & 0.705\\
\cline{2-8}
& FullLora-distractors & 0 &  & 0.069 &  & 0.41 & \\
& LoraQKAtt-distractors & 0.002 &  & 0.09 &  & 0.43 & \\
& FullLora-distill-distractors & 0.001 &  & 0.295 &  & 0.6 & \\
\hline
\multirow{11}{*}{\rotatebox{90}{ RobustQA-Writing }} & Vanilla RAG & 0 & 0.669 & 0.449 & 0.45 & 0.55 & 0.669\\
& FullLora & 0.0 & 0.557 & 0.044 & 0.059 & 0.45 & 0.557\\
& LoraQKAtt & 0.0 & 0.619 & 0.097 & 0.105 & 0.53 & 0.619\\
& LoraAtt & 0 & 0.519 & 0.039 & 0.044 & 0.42 & 0.519\\
& LoraMLP & 0 & 0.56 & 0.043 & 0.064 & 0.43 & 0.56\\
\cline{2-8}
& FullLora-distill & 0 & 0.734 & 0.323 & 0.35 & 0.66 & 0.734\\
& LoraQKAtt-distill & 0 & 0.72 & 0.314 & 0.369 & 0.64 & 0.72\\
& LoraAtt-distill & 0 & 0.727 & 0.339 & 0.36 & 0.65 & 0.727\\
& LoraMLP-distill & 0 & 0.734 & 0.325 & 0.355 & 0.66 & 0.734\\
\cline{2-8}
& FullLora-distractors & 0 &  & 0.047 &  & 0.45 & \\
& LoraQKAtt-distractors & 0.0 &  & 0.076 &  & 0.48 & \\
& FullLora-distill-distractors & 0 &  & 0.286 &  & 0.64 & \\
\hline
\multirow{11}{*}{\rotatebox{90}{ Long-Form Avg }} & Vanilla RAG & 0.0 & 0.61 & 0.42 & 0.42 & 0.52 & 0.61\\
& FullLora & 0.0 & 0.49 & 0.05 & 0.07 & 0.42 & 0.49\\
& LoraQKAtt & 0.0 & 0.58 & 0.12 & 0.13 & 0.49 & 0.58\\
& LoraAtt & 0.0 & 0.47 & 0.05 & 0.06 & 0.41 & 0.47\\
& LoraMLP & 0.0 & 0.49 & 0.05 & 0.07 & 0.41 & 0.49\\
\cline{2-8}
& FullLora-distill & 0.0 & 0.67 & 0.32 & 0.34 & 0.61 & 0.67\\
& LoraQKAtt-distill & 0.0 & 0.66 & 0.33 & 0.36 & 0.59 & 0.66\\
& LoraAtt-distill & 0.0 & 0.66 & 0.33 & 0.34 & 0.6 & 0.66\\
& LoraMLP-distill & 0.0 & 0.67 & 0.32 & 0.34 & 0.61 & 0.67\\
\cline{2-8}
& FullLora-distractors & 0.0 &  & 0.05 &  & 0.41 & \\
& LoraQKAtt-distractors & 0.0 &  & 0.09 &  & 0.45 & \\
& FullLora-distill-distractors & 0.0 &  & 0.29 &  & 0.59 & \\
\hline
\end{tabular}
\caption{RAG Adaptation, Long-Form domain (2). }
\label{tab:rag_adaptation_full_longform2}
\end{table*}

\begin{table*}
\centering
\begin{tabular}{|c|c|cccccc|}
\hline
 & \multirow{2}{*}{\textbf{Adaptation Type}} & \multicolumn{2}{c|}{\textbf{Match}} & \multicolumn{2}{c|}{\textbf{Recall}} & \multicolumn{2}{c|}{\textbf{LLMeval}} \\ \cline{3-8}
 & & \textbf{Llm-1B} & \textbf{Llm-8B} & \textbf{Llm-1B} & \textbf{Llm-8B} & \textbf{Llm-1B} & \textbf{Llm-8B} \\ \hline
\multirow{11}{*}{\rotatebox{90}{ ParaphraseRC }} & Vanilla RAG & 0.378 & 0.625 & 0.481 & 0.605 & 0.49 & 0.625\\
& FullLora & 0.366 & 0.655 & 0.434 & 0.528 & 0.55 & 0.655\\
& LoraQKAtt & 0.366 & 0.649 & 0.434 & 0.541 & 0.53 & 0.649\\
& LoraAtt & 0.373 & 0.654 & 0.437 & 0.525 & 0.55 & 0.654\\
& LoraMLP & 0.367 & 0.654 & 0.428 & 0.524 & 0.55 & 0.654\\
\cline{2-8}
& FullLora-distill & 0.43 & 0.617 & 0.537 & 0.605 & 0.55 & 0.617\\
& LoraQKAtt-distill & 0.42 & 0.625 & 0.524 & 0.609 & 0.53 & 0.625\\
& LoraAtt-distill & 0.43 & 0.616 & 0.537 & 0.605 & 0.54 & 0.616\\
& LoraMLP-distill & 0.432 & 0.62 & 0.539 & 0.607 & 0.55 & 0.62\\
\cline{2-8}
& FullLora-distractors & 0.343 &  & 0.405 &  & 0.53 & \\
& LoraQKAtt-distractors & 0.34 &  & 0.406 &  & 0.51 & \\
& FullLora-distill-distractors & 0.384 &  & 0.494 &  & 0.5 & \\
\hline
\multirow{11}{*}{\rotatebox{90}{ SyllabusQA }} & Vanilla RAG & 0.053 & 0.257 & 0.289 & 0.31 & 0.3 & 0.257\\
& FullLora & 0.092 & 0.346 & 0.149 & 0.178 & 0.33 & 0.346\\
& LoraQKAtt & 0.092 & 0.379 & 0.203 & 0.257 & 0.35 & 0.379\\
& LoraAtt & 0.091 & 0.374 & 0.152 & 0.18 & 0.3 & 0.374\\
& LoraMLP & 0.095 & 0.362 & 0.154 & 0.18 & 0.29 & 0.362\\
\cline{2-8}
& FullLora-distill & 0.08 & 0.366 & 0.347 & 0.342 & 0.35 & 0.366\\
& LoraQKAtt-distill & 0.069 & 0.32 & 0.319 & 0.334 & 0.33 & 0.32\\
& LoraAtt-distill & 0.073 & 0.358 & 0.339 & 0.339 & 0.34 & 0.358\\
& LoraMLP-distill & 0.072 & 0.361 & 0.337 & 0.343 & 0.34 & 0.361\\
\cline{2-8}
& FullLora-distractors & 0.097 &  & 0.167 &  & 0.32 & \\
& LoraQKAtt-distractors & 0.106 &  & 0.208 &  & 0.35 & \\
& FullLora-distill-distractors & 0.103 &  & 0.341 &  & 0.35 & \\
\hline
\multirow{11}{*}{\rotatebox{90}{ TechQA }} & Vanilla RAG & 0.019 & 0.591 & 0.347 & 0.468 & 0.54 & 0.591\\
& FullLora & 0.035 & 0.46 & 0.121 & 0.214 & 0.36 & 0.46\\
& LoraQKAtt & 0.035 & 0.59 & 0.202 & 0.275 & 0.44 & 0.59\\
& LoraAtt & 0.019 & 0.457 & 0.113 & 0.205 & 0.35 & 0.457\\
& LoraMLP & 0.023 & 0.448 & 0.12 & 0.189 & 0.38 & 0.448\\
\cline{2-8}
& FullLora-distill & 0.019 & 0.649 & 0.394 & 0.445 & 0.58 & 0.649\\
& LoraQKAtt-distill & 0.016 & 0.658 & 0.386 & 0.462 & 0.58 & 0.658\\
& LoraAtt-distill & 0.016 & 0.647 & 0.401 & 0.448 & 0.58 & 0.647\\
& LoraMLP-distill & 0.019 & 0.651 & 0.398 & 0.444 & 0.61 & 0.651\\
\cline{2-8}
& FullLora-distractors & 0.018 &  & 0.119 &  & 0.37 & \\
& LoraQKAtt-distractors & 0.039 &  & 0.176 &  & 0.46 & \\
& FullLora-distill-distractors & 0.019 &  & 0.357 &  & 0.54 & \\
\hline
\multirow{11}{*}{\rotatebox{90}{ Context-Critical Avg }} & Vanilla RAG & 0.15 & 0.49 & 0.37 & 0.46 & 0.44 & 0.49\\
& FullLora & 0.16 & 0.49 & 0.23 & 0.31 & 0.41 & 0.49\\
& LoraQKAtt & 0.16 & 0.54 & 0.28 & 0.36 & 0.44 & 0.54\\
& LoraAtt & 0.16 & 0.5 & 0.23 & 0.3 & 0.4 & 0.5\\
& LoraMLP & 0.16 & 0.49 & 0.23 & 0.3 & 0.41 & 0.49\\
\cline{2-8}
& FullLora-distill & 0.18 & 0.54 & 0.43 & 0.46 & 0.49 & 0.54\\
& LoraQKAtt-distill & 0.17 & 0.53 & 0.41 & 0.47 & 0.48 & 0.53\\
& LoraAtt-distill & 0.17 & 0.54 & 0.43 & 0.46 & 0.49 & 0.54\\
& LoraMLP-distill & 0.17 & 0.54 & 0.42 & 0.46 & 0.5 & 0.54\\
\cline{2-8}
& FullLora-distractors & 0.15 &  & 0.23 &  & 0.4 & \\
& LoraQKAtt-distractors & 0.16 &  & 0.26 &  & 0.43 & \\
& FullLora-distill-distractors & 0.17 &  & 0.4 &  & 0.46 & \\
\hline
\end{tabular}
\caption{RAG Adaptation, Context-Critical domain. }
\label{tab:rag_adaptation_full_contextcritical}
\end{table*}

\begin{table*}
\centering
\begin{tabular}{|c|c|cccccc|}
\hline
 & \multirow{2}{*}{\textbf{Adaptation Type}} & \multicolumn{2}{c|}{\textbf{Match}} & \multicolumn{2}{c|}{\textbf{Recall}} & \multicolumn{2}{c|}{\textbf{LLMeval}} \\ \cline{3-8}
 & & \textbf{Llm-1B} & \textbf{Llm-8B} & \textbf{Llm-1B} & \textbf{Llm-8B} & \textbf{Llm-1B} & \textbf{Llm-8B} \\ \hline
\multirow{11}{*}{\rotatebox{90}{ SearchQA }} & Vanilla RAG & 0.431 & 0.745 & 0.457 & 0.726 & 0.53 & 0.745\\
& FullLora & 0.433 & 0.824 & 0.562 & 0.725 & 0.68 & 0.824\\
& LoraQKAtt & 0.433 & 0.797 & 0.46 & 0.7 & 0.59 & 0.797\\
& LoraAtt & 0.471 & 0.832 & 0.498 & 0.734 & 0.62 & 0.832\\
& LoraMLP & 0.521 & 0.826 & 0.549 & 0.727 & 0.67 & 0.826\\
\cline{2-8}
& FullLora-distill & 0.569 & 0.797 & 0.598 & 0.72 & 0.7 & 0.797\\
& LoraQKAtt-distill & 0.563 & 0.795 & 0.593 & 0.748 & 0.69 & 0.795\\
& LoraAtt-distill & 0.57 & 0.79 & 0.602 & 0.73 & 0.69 & 0.79\\
& LoraMLP-distill & 0.574 & 0.791 & 0.603 & 0.72 & 0.7 & 0.791\\
\cline{2-8}
& FullLora-distractors & 0.525 &  & 0.552 &  & 0.67 & \\
& LoraQKAtt-distractors & 0.433 &  & 0.461 &  & 0.59 & \\
& FullLora-distill-distractors & 0.586 &  & 0.618 &  & 0.69 & \\
\hline
\end{tabular}
\caption{RAG Adaptation, Web Search. }
\label{tab:rag_adaptation_full_websearch}
\end{table*}

\begin{table*}
\centering
\begin{tabular}{|c|c|cccccc|}
\hline
 & \multirow{2}{*}{\textbf{Adaptation Type}} & \multicolumn{2}{c|}{\textbf{Match}} & \multicolumn{2}{c|}{\textbf{Recall}} & \multicolumn{2}{c|}{\textbf{LLMeval}} \\ \cline{3-8}
 & & \textbf{Llm-1B} & \textbf{Llm-8B} & \textbf{Llm-1B} & \textbf{Llm-8B} & \textbf{Llm-1B} & \textbf{Llm-8B} \\ \hline
\multirow{11}{*}{\rotatebox{90}{ Avg }} & Vanilla RAG & 0.176 & 0.61 & 0.443 & 0.511 & 0.51 & 0.61\\
& FullLora & 0.19 & 0.575 & 0.252 & 0.306 & 0.5 & 0.58\\
& LoraQKAtt & 0.185 & 0.616 & 0.286 & 0.335 & 0.53 & 0.62\\
& LoraAtt & 0.184 & 0.568 & 0.244 & 0.297 & 0.49 & 0.57\\
& LoraMLP & 0.19 & 0.576 & 0.25 & 0.303 & 0.49 & 0.58\\
\cline{2-8}
& FullLora-distill & 0.219 & 0.657 & 0.45 & 0.483 & 0.6 & 0.66\\
& LoraQKAtt-distill & 0.211 & 0.65 & 0.446 & 0.495 & 0.59 & 0.65\\
& LoraAtt-distill & 0.217 & 0.65 & 0.455 & 0.486 & 0.6 & 0.65\\
& LoraMLP-distill & 0.218 & 0.656 & 0.452 & 0.484 & 0.61 & 0.66\\
\cline{2-8}
& FullLora-distractors & 0.18 &  & 0.244 &  & 0.49 & \\
& LoraQKAtt-distractors & 0.176 &  & 0.264 &  & 0.5 & \\
& FullLora-distill-distractors & 0.21 &  & 0.423 &  & 0.58 & \\
\hline
\end{tabular}
\caption{RAG Adaptation, Average across all datasets.}
\label{tab:rag_adaptation_full_avg}

\end{table*}

\clearpage

\end{document}